\documentclass[10pt,journal,compsoc]{IEEEtran}
\ifCLASSOPTIONcompsoc
  \usepackage[nocompress]{cite}
\else
  \usepackage{cite}
\fi
\ifCLASSINFOpdf
\else
\fi
\usepackage{epsfig}
\usepackage{graphicx}
\usepackage{amsmath}
\usepackage{amssymb}
\usepackage{algorithm}
\usepackage{algorithmic}
\usepackage{multirow}
\usepackage{graphics}
\usepackage{graphicx}
\usepackage{subfigure}
\usepackage{caption2}
\usepackage{soul,color}
\usepackage{color}
\usepackage{rotating}
\usepackage{booktabs}
\usepackage{tabularx}
\usepackage{makecell}
\usepackage{arydshln}

\usepackage[nocompress]{cite}

\begin{document}
%
\title{Sparse Representation based Multi-sensor Image Fusion: A Review}
%
%
%
%

\author{Qiang~Zhang,
~Yi~Liu,
~Rick S.~Blum,~\IEEEmembership{Fellow,~IEEE},
~Jungong~Han,
and~Dacheng~Tao,~\IEEEmembership{Fellow,~IEEE}
\IEEEcompsocitemizethanks{\IEEEcompsocthanksitem Q. Zhang is with the Key Laboratory of Electronic Equipment Structure Design, Ministry of Education, Xidian University, China. He and Y. Liu are also with the Center for Complex Systems, School of Mechano-Electronic Engineering, Xidian University, Xi'an Shaanxi 710071, China. Email: qzhang@xidian.edu.cn, yliu\underline{ }89@stu.xidian.edu.cn.\protect\\

\IEEEcompsocthanksitem R. S. Blum is with Electrical and Computer Engineering Department, Lehigh University, Bethlehem, PA 18015, United States. Email: rb0f@lehigh.edu.
\IEEEcompsocthanksitem J. Han is with Department of Computer and Information Sciences, Northumbria University, Newcastle upon Tyne NE1 8ST, U.K. Email: jungong.han@northumbria.ac.uk. (corresponding author)
\IEEEcompsocthanksitem D.Tao is with the School of Information Technologies in the Faculty of Engineering and Information Technologies at the University of Sydney, J12/318 Cleveland St, Darlington NSW 2008, Australia. Email:dacheng.tao@sydney.edu.au.}
\thanks{Manuscript xxx, 2017}}

\IEEEtitleabstractindextext{%
\begin{abstract}
As a result of several successful applications in computer vision and image processing, sparse representation (SR) has attracted significant attention in multi-sensor image fusion. Unlike the traditional multiscale transforms (MSTs) that presume the basis functions, SR learns an over-complete dictionary from a set of training images for image fusion, and it achieves more stable and meaningful representations of the source images. By doing so, the SR-based fusion methods generally outperform the traditional MST-based image fusion methods in both subjective and objective tests. In addition, they are less susceptible to mis-registration among the source images, thus facilitating the practical applications. This survey paper proposes a systematic review of the SR-based multi-sensor image fusion literature, highlighting the pros and cons of each category of approaches. Specifically, we start by performing a theoretical investigation of the entire system from three key algorithmic aspects, (1) sparse representation models; (2) dictionary learning methods; and (3) activity levels and fusion rules. Subsequently, we show how the existing works address these scientific problems and design the appropriate fusion rules for each application such as multi-focus image fusion and multi-modality (e.g., infrared and visible) image fusion. At last, we carry out some experiments to evaluate the impact of these three algorithmic components on the fusion performance when dealing with different applications. This article is expected to serve as a tutorial and source of reference for researchers preparing to enter the field or who desire to employ the sparse representation theory in other fields.
\end{abstract}

}

\maketitle

\IEEEdisplaynontitleabstractindextext

%
\IEEEpeerreviewmaketitle

\IEEEraisesectionheading{\section{Introduction}\label{sec:introduction}}

\IEEEPARstart{D}{ue} to recent technological advancements, extensive varieties of imaging sensors have been employed in many applications including remote sensing, medical imaging, video surveillance, machine vision and security. Thus, finding a way to most effectively utilize the information captured from these multiple sensors, possibly of different modalities, is of considerable interest. Image fusion provides one versatile solution, where multiple aligned images acquired by different sensors are merged into a composite image. The properly fused image is more informative than any of the individual input images and can thus better interpret the scene \cite{liu2015multi-focus}. As a result, multi-sensor image fusion has always been an active research topic, facilitating a variety of vision-related applications.

	To date, a large number of image fusion algorithms have been proposed \cite{li2017pixel-level,zhang2015multisensor,pertuz2013generation,james2014medical,li2013image}, in which multiscale transform-based (MST) fusion methods are the most popular \cite{li2011performance,Pajares2004A,Zhang1999A}. Traditional MST fusion methods are generally those based on pyramids \cite{Zhang2013Multimodality} and wavelet transforms \cite{Liu2014Region}. Recently developed fusion methods can be considered as their variations and extensions employing multiscale geometric analysis (MGA) tools, such as the Curvelet Transform \cite{guo2012multifocus}, the Shearlet Transform \cite{wang2014multi-modal} and the nonsubsampled Contourlet Transform (NSCT) \cite{upla2015an}. Thorough reviews on such methods can be found in \cite{li2017pixel-level,li2011performance}.

	Sparse representation (SR) \cite{wright2009robust} has recently drawn significant interest in computer vision and image processing due to its enhanced performance in many applications, such as face recognition \cite{wright2009robust}, action recognition \cite{guha2012learning}, and object tracking \cite{yuan2012visual}. The main idea of SR theory lies in the fact that an image signal can be represented as a linear combination of the fewest possible atoms or transform basis primitives in an over-complete dictionary. Sparsity means that only a small number of atoms are required to accurately reconstruct a signal, i.e., the coefficients become sparse. Over-completeness indicates that the number of atoms in the dictionary is larger than the dimension of the signal. Thus, a sufficient number of atoms in an over-complete dictionary permit an accurate sparse representation of signals \cite{yang2010multifocus}.

	Not surprisingly, SR has also attracted significant attention in the research field of image fusion \cite{yang2010multifocus, Liu2015A,zhang2016an,wei2015hyperspectral}. Similar to the traditional MST-based image fusion methods, most of the SR-based image fusion methods also belong to the transform-domain-based techniques\footnote{As discussed later, parts of the SR-based fusion methods belong to the spatial-domain-based methods.}. However, there are two main differences between the SR-based and the traditional MST-based fusion methods \cite{yang2010multifocus,Liu2015A}.
 \begin{enumerate}
    \item  The traditional MSTs usually fix their basis functions in advance for image analysis and fusion. Due to the limitations of predefined basis functions, some significant features (e.g., edges) of source images may not be well expressed and extracted, thereby dramatically degrading the performance of fusion. In contrast, SR generally \emph{learns} an over-complete dictionary from a set of training images for image fusion, which captures intrinsic data-driven image representations tending to be domain agnostic. The over-complete dictionary contains richer basis atoms allowing more meaningful and stable representations of source images. By doing so, SR-based fusion methods generally outperform the traditional MST-based image fusion methods in both subjective and objective tests.
    \item   The traditional MST-based fusion methods are implemented in a multiscale manner, where the selection of the MST decomposition level becomes thereby crucial and tricky. To ensure spatial details can be extracted from the source images, the decomposition level is often set too large. In this case, one coefficient in the low-pass band has a great impact on a large set of pixels in the fused image. Accordingly, an error in the low-pass sub-band (mainly caused by noise or mis-registration between the source images) will lead to serious artificial effects \cite{Liu2015A}. The fusion of the high-pass sub-band coefficients is also sensitive to noise and mis-registration in this case. Consequently, the MST-based fusion methods are generally sensitive to mis-registration, impending their usage in the practical applications where a perfect spatial alignment of different source images is unachievable. In contrast, the SR-based fusion methods are generally implemented in a patch-based way. More specifically, the source images are first divided into a number of patches of the same size, and the fusion is carried out at the patch level. Moreover, in order to reduce block artifacts and improve the robustness against mis-registration, a sliding window with a step length equal to a fixed number of pixels (e.g., one pixel) is often used in the SR-based fusion methods. In other words, these patches overlap by a fixed number of pixels along the horizontal and vertical directions. Generally, SR-based fusion methods are more robust to mis-registration than MST-based ones.
  \end{enumerate}

\subsection{SR image fusion in a nutshel}
Since Yang and Li \cite{yang2010multifocus} took the first step in applying the SR theory to the image fusion field, a number of SR-based image fusion methods have been proposed. As shown in Fig. 1, the growing appeal of this research area can be observed from the steady increase in the number of scientific papers published in academic journals and magazines since 2010.

\begin{figure}[h]
  \centering
  \includegraphics[width=0.95\linewidth]{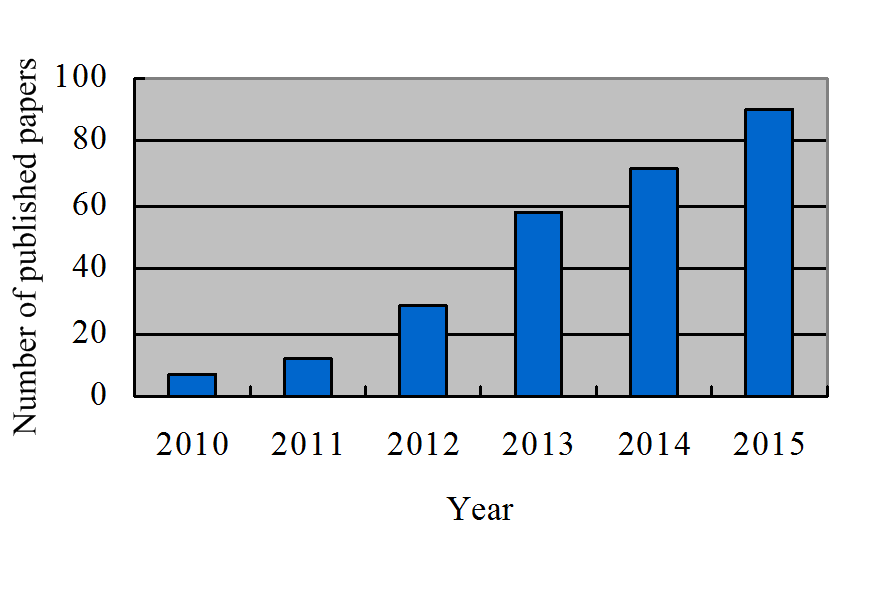}\\
  \caption{Numbers of publications on SR-based fusion methods, obtained from the Web of Science indexing service.}\label{1}
\end{figure}
The basic idea behind SR-based image fusion is that image signals can be represented as a linear combination of a \lq\lq few\rq\rq\  atoms from a pre-learned dictionary, and the sparse coefficients describe the salient features of the source images. As shown in Fig. 2, the main steps in most SR-based image fusion methods include: (a) segment the source images into some overlapping patches and rewrite each of these patches as a vector; (b) perform sparse representation on the source image patches using pre-defined or learned dictionaries; (c) combine the sparse representations by some fusion rules; (d) reconstruct the fused images from their sparse representations.

\begin{figure*}[!t]
  \centering
  \includegraphics[width=0.95\textwidth]{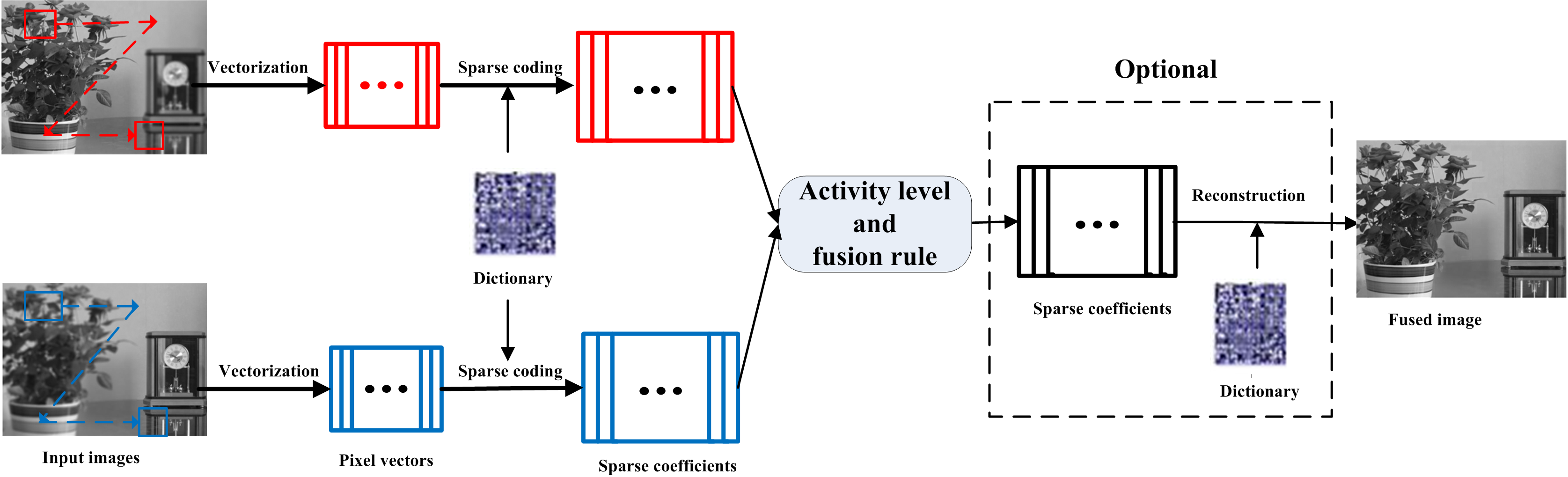}\\
  \caption{Diagram of the SR-based image fusion method. (Credit to [2])}\label{2}
\end{figure*}

	The dictionaries employed in these methods may be directly obtained from some fixed (e.g., DCT and Wavelet) basis \cite{yang2010multifocus}. They can also be learned from a set of auxiliary images (\emph{global trained dictionary}) \cite{liu2015simultaneous} or from the input images themselves (\emph{adaptively trained dictionary}) \cite{nejati2015multi-focus} using some learning methods, such as K-SVD \cite{Aharon2006}. Sometimes, a pair of coupled dictionaries are even simultaneously learned from a high-spatial-resolution image and its spatially-degraded version. Using the coupled dictionaries allows to produce a fused image with higher spatial-resolution \cite{yin2013simultaneous,guo2014an}.

Different sparse representation models have been used in image fusion methods. They include: (1) the traditional SR model \cite{wright2009robust} in which the sparsity constraint (using $l_0$-norm or $l_1$-norm) is performed on the representation coefficients; (2) the non-negative SR model \cite{wang2014fusion} in which the sparsity and non-negativity constraints are jointly imposed on the representation coefficients; (3) the robust SR model \cite{zhang2016robust} in which the sparsity constraint is imposed on the reconstruction errors as well as on the representation coefficients; (4) the group-sparsity SR model \cite{li2012group-sparse} in which the nonzero representation coefficients are forced to occur in clusters (called group-sparsity) rather than appear randomly; (5) the joint-sparse representation (JSR) model \cite{zhang2013dictionary} which indicates that different signals from various sensors of the same scene form an ensemble. All signals in one ensemble have a common sparse component, and each employs an individual sparse component.

	When fusing the source image patches, the $l_1$- or $l_2$-norm of the representation coefficients \cite{yang2010multifocus} is generally used. It could possibly benefit from other information to calculate the activity level \cite{Zhang1999A}, which measures the information contained in these representation coefficients that is deemed useful during the fusion. Statistical characteristics, such as the sparseness level \cite{wang2014fusion} of their representation coefficients, might also be employed to determine the activity level during the fusion. The energy of the sparse reconstruction errors \cite{zhang2016robust} has been used to determine the activity level when fusing multi-focus images. With an activity calculation defined, a maximum-selecting or a weighted-averaging fusion rule can be employed to directly combine source image patches or indirectly combine representation coefficients of the source image patches \cite{Zhang1999A}. If the representation coefficients are to be combined, the fused image is reconstructed using the pre-learned dictionary and the combined representation coefficients (called the transform-domain fusion method) \cite{wang2014fusion,li2012group-sparse,zhang2013dictionary,kim2016joint}. Otherwise, the fused image can be directly obtained from the source image patches according to their activity level (called the spatial-domain fusion method) \cite{nejati2015multi-focus,zhang2016robust}. The preferred approach depends on the specific intended applications (e.g., fusion of multi-focus images or multi-modality images).

	Based on the above analysis, in this paper we will review sparse representation (SR) image fusion methods from the following four key aspects: (1) sparse representation models; (2) dictionary learning methods; (3) activity levels and fusion rules; and finally, (4) applications to multi-focus images and multi-modality (e.g., infrared and visible) image fusion.

\subsection{Why this survey?}
As pointed out previously, multi-sensor image fusion has always been a hot research topic in the area of image processing, and a considerable number of publications emerge every year. The early reviews \cite{li2017pixel-level,james2014medical,Zhang1999A,vivone2015a,loncan2015hyperspectral} that focus mainly on traditional MST-based \cite{li2017pixel-level,Zhang1999A} or spatial-domain-based (e.g., patches) fusion methods are outdated as they missed out on important recent advances, such as SR-based image fusion methods. In addition, most of them are only limited to one single application of image fusion, such as multi-focus \cite{Zhang1999A}, medical \cite{james2014medical} or remote sensing image fusion \cite{vivone2015a,loncan2015hyperspectral}. On the other hand, in this paper, we will thoroughly discuss the SR-based fusion methods as well as their applications to fusion of both multi-focus and multi-modality images. Recently, some review papers have also appeared on sparse representation theory \cite{wright2010sparse,zhang2015a} with the aim to explain the mathematical and theoretical aspects of SR models, but they do not particularly discuss image fusion problems. To the best of our knowledge, there are no previous papers where SR-based fusion methods are reviewed and evaluated. Therefore, it is desirable to put a thorough survey concerning SR-based image fusion in place, which may be useful to a variety of audience, ranging from image fusion learners intended to quickly grasp the current progress in this research area as a whole, to image fusion practitioners interested in applying SR methods to their own problems.

\begin{figure*}[!t]
  \centering
  \includegraphics[width=1\textwidth]{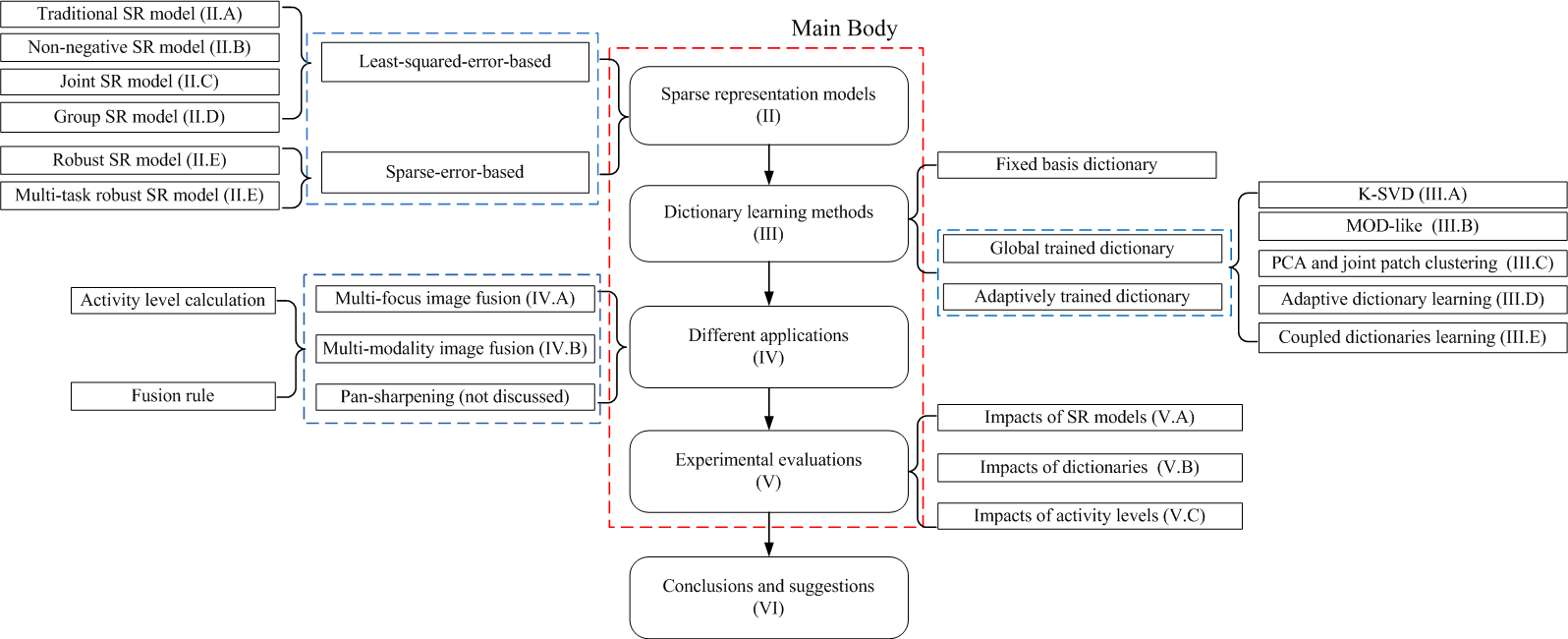}\\
  \caption{Organization of this paper.}\label{3}
\end{figure*}

\subsection{Paper outline}
The rest of this paper is organized as follows. The available SR models are thoroughly reviewed in Section II. In Section III, dictionary learning methods are surveyed. In Section IV, the activity level calculations and fusion rules exploited in the literature with different applications are discussed. In Section V, the impact of the choice of the components presented in Sections II, III and IV on the fusion performance is examined. Finally, conclusions and suggestions for future work are provided in Section VI. Fig. 3 summarizes the structure of this paper.

\subsection{Notations}
	We assume that the reader has some basic knowledge of linear algebra and optimization theories. Throughout the paper, a vector is denoted by a low-case letter. A matrix is denoted by a capital letter. All the elements in a vector or a matrix are real-valued. Given a vector \emph{x} and a matrix \emph{X}, some notations related to them used in this paper are listed in Table 1.

\begin{table}[h]
\caption{List of vector and matrix related notations}
\vspace*{-0pt}
\begin{center}
\begin{tabular}{c|p{7.1 cm}}
\toprule
Symbols & Definition \\
\midrule
$x(i)$ & the $i$-th entry of the vector $x$ \\
$X(i,j)$ & the $(i,j)$-th entry of the matrix $X$\\
${\left\| x \right\|_0}$ & $l_0$-norm of the vector $x$, i.e., the number of nonzero entries in the vector $x$\\
${\left\| x \right\|_1}$ & $l_1$-norm of the vector $x$, ${\left\| x \right\|_1} = \sum\nolimits_i {\left| {x(i)} \right|} $\\
${\left\| x \right\|_2}$ & $l_2$-norm of the vector $x$, ${\left\| x \right\|_2} = \sqrt {\sum\nolimits_i {{x^2}(i)} } $\\
${\left\| X \right\|_0}$ & $l_0$-norm of the matrix $X$, i.e., the number of nonzero entries in the matrix $X$\\
${\left\| X \right\|_1}$ & $l_1$-norm of the matrix $X$, ${\left\| X \right\|_1} = \sum\nolimits_{i,j} {\left| {X(i,j)} \right|} $\\
${\left\| X \right\|_F}$ & Frobenius -norm of the matrix $X$, ${\left\| X \right\|_F} = \sqrt {\sum\nolimits_{i,j} {{X^2}(i,j)} } $\\
${\left\| X \right\|_{2,1}}$ & $l_{2,1}$-norm of the matrix $X$, ${\left\| X \right\|_{2,1}} = \sum\nolimits_j {\sqrt {\sum\nolimits_i {{X^2}(i,j)} } } $\\
${\left(  \cdot  \right)^T}$ & transpose of a vector or a matrix\\
${X^\dag }$ & pseudo inverse of the matrix X\\
\bottomrule
\end{tabular}
\end{center}
\label{Tab:notation}

\end{table}

\section{SPARSE REPRESENTATION MODELS}
Since the traditional SR model \cite{wright2009robust} was first applied to multi-sensor image fusion, many of its extensions have also been applied to image fusion. For example, a non-negative sparse representation (NNSR) model was introduced for image fusion in \cite{wang2014fusion}. Unlike the traditional SR model that just imposes the sparsity constraint on the representation coefficients, the NNSR model imposes the joint sparsity and non-negativity constraints on the representation coefficients. From the image patch encoding point of view, the interpretation of NNSR model is more intuitive than the traditional SR model.

	Assuming the imaging sensors observe the same scene, the source images captured by these sensors are expected to possess common (or redundant) and complementary (distinct) features. Such ideas map well into the joint sparse representation (JSR) model \cite{zhang2013dictionary}, in which all the each sensor image from the same ensemble is automatically decomposed into a common component that can be shared by all the images and an innovation component that describes individual differebces. As a result, the JSR model attracts more attention in image fusion, especially in multi-modality image fusion.

	In \cite{zhang2016robust}, a robust sparse representation (RSR) model was introduced to extract the detailed information in a set of multi-focus input images. The RSR model  replaces the conventional least-squared reconstruction error with a so-called sparse reconstruction error. By using RSR, any multi-focus image can be decomposed into a fully-defocus image and a sparse but detailed image denoted by the sparse reconstruction error. Distinct from traditional SR-based fusion methods, the reconstruction errors are employed instead of the usual sparse representation coefficients to guide the fusion process. Superiority over the latter SR-based methods is verified in the experimental results.

	In this section, we will review some SR models that have been applied in multi-sensor image fusion. We will start by introducing some specific concepts related to sparse representation, so that the reader can understand the basic concepts associated with this theory. Then we will extend these concepts to some more complex representation models.
\subsection{Sparse representation (SR) model}
The sparse representation model relies on the assumption that many important signals can be represented or approximately represented as a linear combination of a \lq\lq few\rq\rq\  atoms from a redundant dictionary \cite{Liu2015A,nejati2015multi-focus}. That is, given such a redundant dictionary $D \in {R^{n \times M}}$ ($n < M$
) containing \emph{M} prototype \emph{n}-dimensional signals that are referred to as atoms formed by the columns of the matrix \emph{M}, a signal $y \in {R^n}$ can be expressed as $y = Dx$ or $y \approx Dx$. The vector $x \in {R^M}$ contains the coefficients that represent the signal \emph{y} in terms of the dictionary \emph{D}. As the dictionary is redundant, the vector \emph{x} is not unique. Thus, the SR model was proposed as a method for determining the solution vector \emph{x} with the fewest non-zero components \cite{nejati2015multi-focus}. Mathematically, this can be achieved exactly assuming negligible noise or inexactly considering noise by solving the optimization problem
\begin{equation}\label{1}
  \mathop {\min }\limits_x {\left\| x \right\|_0}{\rm{     }}\ s.t.\ {\rm{ }}y = Dx,
\end{equation}
or
\begin{equation}\label{2}
  \mathop {\min }\limits_x {\left\| x \right\|_0}{\rm{     }}\ s.t.\ {\rm{ }}\left\| {y - Dx} \right\|_2^2 \le \varepsilon.
\end{equation}
The optimization of the above formulas is NP-hard and thus requires approximate techniques, such as the matching pursuit (MP) \cite{mallat1993matching}, orthogonal matching pursuit (OMP) \cite{bruckstein2009from} or simultaneous OMP (SOMP) \cite{tropp2006algorithms} algorithms to obtain solutions with low complexity.

Based on recent developments in SR and compressed sensing, the non-convex ${l_0}$-minimization problems in (1) and (2) can be relaxed to obtain the convex ${l_1}$-minimization problems \cite{wright2009robust,candes2011robust} in
\begin{equation}\label{3}
  \mathop {\min }\limits_x {\left\| x \right\|_1}{\rm{     }}\ s.t.\ {\rm{ }}y = Dx,
\end{equation}
and
\begin{equation}\label{4}
  \mathop {\min }\limits_x {\left\| x \right\|_1}{\rm{     }}\ s.t.\ {\rm{ }}\left\| {y - Dx} \right\|_2^2 \le \varepsilon
\end{equation}
Solutions can be obtained by using linear programming methods \cite{wright2009robust,donoho2008fast}.
\subsection{Non-negative sparse representation (NNSR) model}
Considering that properly scaled black and while images can be interpreted as images with positive entries, the authors of \cite{wang2014fusion} introduced a non-negative sparse representation (NNSR) model and applied it to the fusion of infrared and visible light images. Different from the traditional SR model which only emphasizes the sparsity constraint using $l_0$-norm or $l_1$-norm, NNSR jointly imposes the sparsity and non-negativity constraints on the representation coefficients. It can also be seen as an extension of the traditional non-negative matrix factorization \cite{lee1999learning} which adds a sparsity inducing penalty.

Let $Y = [{y_1},{y_2},...,{y_N}]$ be an observed non-negative data matrix\footnote{Here, a matrix $D = [{d_{i,j}}]$ is called non-negative if each of its elements ${d_{i,j}}$ is non-negative. For simplicity, a non-negative matrix \emph{D} is denoted by $D \ge 0$} of size $n \times N$ representing a set of \emph{N} source image patches, each column of which is a data vector (i.e., an image patch) ${y_i} \in {R^n}$. Then, given a dictionary $D \in {R^{n \times M}}$ with \emph{M} non-negative prototype atoms, the NNSR model coefficients can be obtained from
\begin{equation}\label{5}
\begin{split}
  \mathop {\min }\limits_{{x_i}} \sum\limits_{i = 1}^N {\left( {\frac{1}{2}\left\| {{y_i} - D{x_i}} \right\|_2^2 + \lambda {{\left\| {{x_i}} \right\|}_1}} \right)}\\
  s.t.\ {\rm{   }}D \ge 0,{\rm{  }}{x_i} \ge 0,i = 1,2,...,N
  \end{split},
\end{equation}
where ${x_i} \in {R^M}$ denotes the representation coefficients of the data ${y_i}$. Here, owing to the non-negativity, the $l_1$-norm of the vector ${x_i}$ is also calculated as the sum of the components in the vector ${x_i}$. $\lambda $ refers to the regularization parameter. When $\lambda  = 0$, NNSR is reduced to the non-negative matrix factorization. This problem can be simply and efficiently solved by the non-negative sparse coding algorithm \cite{dong2016hyperspectral}.

Similar to the traditional SR model, NNSR can also encode the source images efficiently by using a few \lq\lq active\rq\rq\  components. In contrast, the non-negativity constraint makes the representation purely additive (allowing no subtractions), thus enabling NNSR to achieve an easy or intuitive interpretation of the encodings of the source images \cite{wang2014fusion}.

\subsection{Joint sparse representation (JSR) model and a generalized version}
The term \lq\lq Joint Sparsity\rq\rq, that is, the common sparsity of the entire signal ensemble, was first introduced in \cite{Baron2012Distributed}. Three joint sparsity models (JSMs) for different situations were presented, \textbf{JSM-1} (sparse common component + innovations), \textbf{JSM-2} (common sparse supports) and \textbf{JSM-3} (non-sparse common +sparse innovations). When different imaging sensors observe the same scene, the source images captured by the sensors are generally expected to possess both \lq\lq common (or correlated)\rq\rq\  and \lq\lq innovation (or complementary)\rq\rq\  information. Accordingly, it is not surprising that JSM-1 has been shown to be more suitable for many image fusion applications, especially for the fusion of multi-modality images \cite{zhang2013dictionary}, when compared with JSM-2 and JSM-3.

In the JSM-1 (or JSM\footnote{In the remaining discussion, the symbol \lq\lq JSM\rq\rq\ denotes the JSM-1 model for simplicity unless expressly specified otherwise.}) model, all signals share a common component while each individual signal contains an innovation component. Let ${Y_k} \in {R^{n \times L}}$ ($k = 1,2,...,K$) denote the \emph{L} signals of dimension n from the \emph{k}-th sensor which can be represented using \cite{zhang2013dictionary}
\begin{equation}\label{6}
  {Y_k} = {Y^C} + Y_k^U = D{X^C} + DX_k^U,k = 1,2,...,K,
\end{equation}
where ${Y^C} = D{X^C}$ denotes the common component for all signals, and $Y_k^U = DX_k^U$ denotes the innovation component for the \emph{k}-th individual signal. $D \in {R^{n \times M}}$ ($n < M$) is an over-complete dictionary. ${X^C}$ and $X_k^U \in {R^{M \times L}}$ are the sparse coefficient matrices for the common and innovation components, respectively.

\begin{figure*}[!t]
  \centering
  \includegraphics[width=0.95\linewidth]{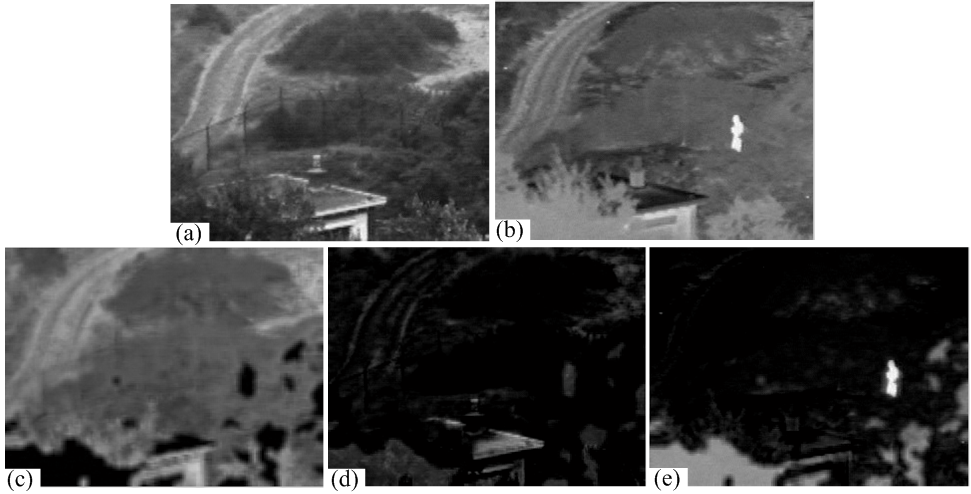}\\
  \caption{ Illustration of the common and innovation information obtained by using the JSR model. (a) and (b) test images captured by two different sensors; (c) The common component between the two test images; (d) and (e) The innovation components of the test images in (a) and (b), respectively.}\label{fig 4}
\end{figure*}

Let
\begin{equation}\label{7}
  Y = \left[ {\begin{array}{*{20}{c}}
   {{Y_1}}  \\
    \vdots   \\
   {{Y_K}}  \\
\end{array}} \right] \in {R^{nK \times L}},
\end{equation}
\begin{equation}\label{8}
  \underline D  = \left[ {\begin{array}{*{20}{c}}
   D & D & {\bf{0}} &  \cdots  & {\bf{0}}  \\
   D & {\bf{0}} & D &  \cdots  & {\bf{0}}  \\
    \vdots  &  \vdots  &  \vdots  &  \ddots  &  \vdots   \\
   D & {\bf{0}} & {\bf{0}} &  \cdots  & D  \\
\end{array}} \right] \in {R^{nK \times (K + 1)M}},
\end{equation}
\begin{equation}\label{9}
  X = \left[ {\begin{array}{*{20}{c}}
   {{X^C}}  \\
   {X_1^U}  \\
    \vdots   \\
   {X_K^U}  \\
\end{array}} \right] \in {R^{(K + 1)M \times L}},
\end{equation}
where ${\bf{0}} \in {R^{n \times M}}$ is a matrix of zeros. Under the assumed sparseness, the coefficients of JSM model can be computed using \cite{zhang2013dictionary,yin2011multimodal,yu2011image}
\begin{equation}\label{10}
  \mathop {\min }\limits_X {\left\| X \right\|_0}{\rm{     }}\ s.t.\ {\rm{  }}\left\| {Y - \underline D X} \right\|_F^2 \le \varepsilon,
\end{equation}
where $\varepsilon  \ge 0$ is the error tolerance. Similar to solving (3) in the traditional SR model, the joint sparse coefficient matrix \emph{X} of the JSM model in (10) can be obtained by using the previously discussed sparse approximation algorithms (e.g., the OMP algorithm \cite{bruckstein2009from}). Fig.\ref{fig 4} illustrates the common and complementary information obtained by using the JSR model\footnote{The test images in Fig. 4, Fig. 10, and Fig. 12 are downloaded from www.imagefusion.org.}, where Fig.\ref{fig 4} (c) contains the common background information acquired by the two sensors, while Fig.\ref{fig 4} (d) and (e) contain the complementary information between the two source images. Especially, the man behind the tree captured by the infrared imaging sensor is clearly displayed in Fig.\ref{fig 4} (e).

Considering that the subspace spanned by the innovation component might not be the same as the subspace spanned by the common component, Zhang et al., \cite{zhang2013dictionary} presented \emph{a generalized version of the JSM model}. In the generalized JSM model, the signals from one ensemble are assumed to depend on two dictionaries, i.e. the common dictionary ${D^C} \in {R^{n \times M}}$ and the innovation dictionary ${D^U} \in {R^{n \times M}}$, instead of a single dictionary as in the JSM model. Accordingly, (6) and the dictionary matrix $\underline D$ in (8) are extended in the generalized JSM model \cite{zhang2013dictionary}, respectively to
\begin{equation}\label{11}
  {Y_k} = {Y^C} + Y_k^U = {D^C}{X^C} + {D^U}X_k^U,k = 1,2,...,K,
\end{equation}
\begin{equation}\label{12}
  \underline D  = \left[ {\begin{array}{*{20}{c}}
   {{D^C}} & {{D^U}} & {\bf{0}} &  \cdots  & {\bf{0}}  \\
   {{D^C}} & {\bf{0}} & {{D^U}} &  \cdots  & {\bf{0}}  \\
    \vdots  &  \vdots  &  \vdots  &  \ddots  &  \vdots   \\
   {{D^C}} & {\bf{0}} & {\bf{0}} &  \cdots  & {{D^U}}  \\
\end{array}} \right] \in {R^{nK \times (K + 1)M}}.
\end{equation}

According to (10), the generalized JSM model can be solved by using the same methods as those for the traditional SR and JSM models. In \cite{zhang2013dictionary}, the generalized JSM model is shown to be sometimes superior to the JSM model in terms of the ability to extract detailed information from the resulting image representations but with little extra computational complexity.

\subsection{Group sparse representation model}
Most of the existing SR models mentioned previously assume that the non-zero coefficients appear randomly, and do not consider the intrinsic structure of the signals. For that, Li, \emph{et al.}, introduced a group sparse representation (GSR) model \cite{li2012group-sparse}, in which the cluster structure sparsity prior is incorporated and the non-zero elements are forced to occur in clusters (called group-sparsity), rather than appear randomly.

Let $G = \left\{ {{G_1},{G_2},...,{G_g}} \right\}$ be a partition of the index set $\left\{ {1,2,...,M} \right\}$, where \emph{g} is the number of groups. Given a dictionary $D = \left[ {{D_{{G_1}}},{D_{{G_2}}},...,{D_{{G_g}}}} \right] \in {R^{n \times M}}$  where ${D_{{G_i}}}$ denotes the sub-dictionary with columns identical to \emph{D} in group ${G_i}$, any signal $y \in {R^n}$ can be represented as [29]
\begin{equation}\label{13}
  y = Dx = \left[ {{D_{{G_1}}},{D_{{G_2}}},...,{D_{{G_g}}}} \right]{\left[ {x_{{G_1}}^T,x_{{G_2}}^T,...,x_{{G_g}}^T} \right]^T},
\end{equation}
where $x = {\left[ {x_{{G_1}}^T,x_{{G_2}}^T,...,x_{{G_g}}^T} \right]^T} \in {R^M}$ denotes the representation coefficients, and ${x_{{G_i}}}$ ($i=1,2,...,g$) are the representation coefficients with respect of the sub-dictionary ${D_{{G_2}}}$. In the GSR model, the sparse representation coefficients are found from
\begin{equation}\label{14}
  \mathop {\min }\limits_x {\left\| x \right\|_{2,0}}{\rm{     }}\ s.t.\ {\rm{ }}y = Dx{\rm{   \ or\    }}\left\| {y - Dx} \right\|_2^2 \le \varepsilon,
\end{equation}
where ${\left\| x \right\|_{2,0}} = \sum\limits_{i = 1}^g {I\left( {{{\left\| {{x_{{G_i}}}} \right\|}_2}} \right)} $ , and $I\left(  \cdot  \right)$ is an indicator function, i.e.,
\begin{equation}\label{15}
I\left( {{{\left\| {{x_{{G_i}}}} \right\|}_2}} \right) = \left\{ {\begin{array}{*{20}{c}}
   {1,} & {{\rm{if}}{{\left\| {{x_{{G_i}}}} \right\|}_2} > 0}  \\
   {0,} & {{\rm{otherwise}}}  \\
\end{array}} \right..
\end{equation}

Similarly, the non-convex ${l_{2,0}}$-minimization optimization problem in (14) can be relaxed by solving the following convex ${l_{2,1}}$-minimization problem in (16)
\begin{equation}\label{16}
  \mathop {\min }\limits_x {\left\| x \right\|_{2,1}}{\rm{     }}\ s.t.\ {\rm{ }}y = Dx{\rm{   \ or\   }}\left\| {y - Dx} \right\|_2^2 \le \varepsilon,
\end{equation}
where ${\left\| x \right\|_{2,1}}{\rm{  = }}\sum\limits_{i = 1}^g {{{\left\| {{x_{{G_i}}}} \right\|}_2}}$. The GSR model can be effectively solved via the Group Orthogonal Matching Pursuit (GOMP) algorithm \cite{majumdar2009fast}.

	Fig. 5 illustrates the representation coefficients obtained by using the SR model and the GSR model. In the GSR model, a dictionary containing 8 sub-dictionaries (i.e., $g=8$ in (13)) is employed. As shown in Fig. 5(b) and (d), the coefficients obtained by using SR model are sparsely and randomly distributed along the entire horizontal axis. In contrast, the coefficients obtained by using the GSR model are just sparsely located at a few segments along the horizontal axis as shown in Fig. 5(c) and (e). This demonstrates that each local patch can be well reconstructed by using only a few sub-dictionaries, instead of a few random dictionary atoms, in the GSR model.

\begin{figure}[ht]
  \centering
  \includegraphics[width=0.95\linewidth]{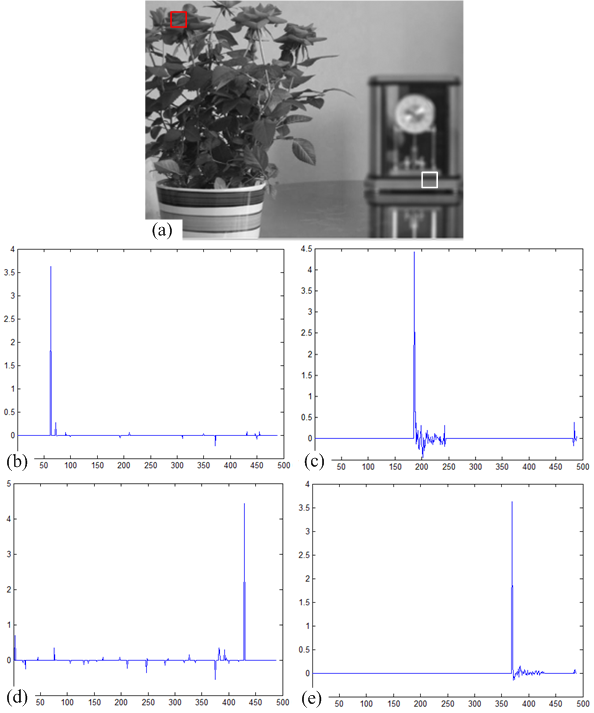}\\
  \caption{Illustration of GSR coefficients. (a) Test image; (b) and (c) SR coefficients and GSR coefficients for the red rectangle patch in (a), respectively; (d) and (e) SR coefficients and GSR coefficients for the white rectangle patch in (a), respectively.}\label{5}
\end{figure}

\subsection{Robust sparse representation (RSR) model and a multi-task version}
As discussed previously, the traditional SR, NNSR, JSR and GSR models are seen to impose either an ${l_0}$-norm or $l_1$-norm minimization on the representation coefficients to achieve a sparse representation of a signal, while imposing an $l_2$-norm minimization on the reconstruction errors (e.g., the component $\frac{1}{2}\left\| {{y_i} - D{x_i}} \right\|_2^2$ in (5))\footnote{In fact, the problems in (3) and (4) are equivalent to the following problem: $\mathop {\min }\limits_x \frac{1}{2}\left\| {y - Dx} \right\|_2^2{\rm{ + }}\lambda {\left\| x \right\|_1}$. Thus, the traditional SR model also imposes an ${l_2}$-norm minimization on the reconstruction errors.}. These approaches work well for signals with small levels of Gaussian noise. However, if the signal contains non-Gaussian noise or is corrupted by sparse but strong \lq\lq outliers\rq\rq, it may not be possible to achieve a satisfactory result \cite{wright2009robust}.

In \cite{zhang2016robust}, Zhang and Levine presented a robust sparse representation (RSR) model by imposing sparse constraints on the reconstruction errors as well as on the representation coefficients. More specifically, let $Y = [{y_1},{y_2},...,{y_N}]$ be an observed data matrix of size $n \times N$, each column of which is a data vector ${y_i} \in {R^n}$. Further, suppose the observed data \emph{Y} is partially corrupted by errors or noise $E \in {R^{n \times N}}$. Then, given a dictionary $D \in {R^{n \times M}}$ with \emph{M} prototype atoms, the coefficients of the RSR model are assumed to follow \cite{zhang2016robust}
\begin{equation}\label{17}
  \mathop {\min }\limits_{X,E} {\rm{  }}{\left\| X \right\|_1} + \lambda {\left\| E \right\|_{2,1}}{\rm{     }}\ s.t.\ {\rm{    }}Y = DX + E,
\end{equation}
where the matrix $X \in {R^{M \times N}}$ denotes the sought after matrix of coefficients, and each of its columns ${x_i} \in {R^M}$ denotes the sparse coefficient vector for the data ${y_i}$. $\lambda  > 0$ is a parameter and is used to balance the effects of the two components in (17). The optimization problem in (17) is convex and can be solved by various methods. In \cite{zhang2016robust}, the authors used the linearized alternating direction method with adaptive penalty (LADMAP) \cite{Lin2011Linearized,zhang2013learning} to solve this problem because of its high efficiency.

Here, we perform an experiment to demonstrate the robustness of the RSR model to non-Gaussian noise or sparse \lq\lq outliers\rq\rq. Similar to \cite{wright2009robust}, we select half of the images in the Extended Yale B database for training and the rest for testing. In the experiment, the pixel intensities of the original images are used as features and stacked as columns of the dictionary matrix \emph{D} and the data matrix \emph{Y}. Then the representation coefficient matrix \emph{X} and reconstruction matrix \emph{E} are obtained by solving (17).

As shown in Fig. 6, the images reconstructed by the RSR model are superior to those reconstructed by the traditional SR model. For example, there are some ghosts near the eye regions labeled by a green rectangle in Fig. 6(b1) reconstructed using the traditional SR model. This phenomenon looks more severe in Fig. 6(b2). In contrast, these ghosts are greatly reduced in the images reconstructed by the RSR model, as shown in Fig. 6(c1) and 6(c2). This also demonstrates that the RSR model is more robust to non-Gaussian noise or sparse \lq\lq outliers\rq\rq\  than the traditional SR model.

\begin{figure*}[!t]
  \centering
  \includegraphics[width=0.95\linewidth]{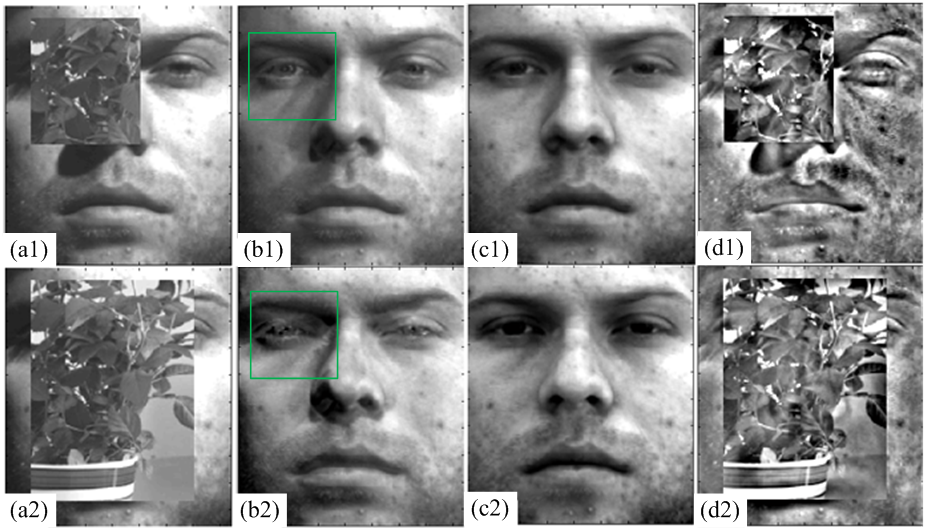}\\
  \caption{Reconstructed results for images with occlusions. (a1) and (a2) are occluded test images of the first subject in the Extended Yale B database with 23\%\ and 61\%\ occlusion, respectively; (b1) and (b2) are reconstructed images using the dictionary atoms from the first subject and their corresponding SR coefficients for (a1) and (a2), respectively; (c1) and (c2) are reconstructed images using the dictionary atoms from the first subject and their corresponding RSR coefficients for (a1) and (a2), respectively; (d1) and (d2) indicate the RSR reconstruction errors for (a1) and (a2), respectively.}\label{6}
\end{figure*}

In order to effectively extract and utilize multiple features for each local image patch during the fusion process, Zhang and Levine generalized the RSR model to multi-task sparsity pursuit and presented a multi-task RSR (MRSR) model \cite{zhang2016robust}. In MRSR, the multi-task sparsity pursuit is achieved by enforcing a joint sparsity constraint on the reconstruction errors across all the tasks.

Let ${Y_k} = \left[ {{y_{k,1}},{y_{k,2}},...,{y_{k,N}}} \right] \in {R^{{n_k} \times N}}$ ($k = 1,2,...,K$) consist of \emph{K} feature matrices for \emph{K} different types of features. The vector ${y_{k,i}} \in {R^{{n_k}}}$ denotes the \emph{k}-th type of feature of dimension ${n_k}$ for the \emph{i}-th image patch. Correspondingly, the columns ${y_{k,i}} \in {R^{{n_k}}}$ ($k = 1,2,...,K$) in these matrices with the same index \emph{i} and different \emph{k} denote different types of features for the same \emph{i}-th image patch. \emph{N} denotes the total number of patches in the image to be considered. Then the MRSR coefficients are assumed to satisfy \cite{zhang2016robust}:
\begin{equation}\label{18}
\begin{split}
  \mathop {\min }\limits_{{X_k},{E_k}} \sum\limits_{k = 1}^K {{{\left\| {{X_k}} \right\|}_1} + \lambda {{\left\| E \right\|}_{2,1}}} {\rm{       }}\\
  s.t.\ {\rm{  }}{Y_k} = {D_k}{X_k} + {E_k}{\rm{   }},\ k = 1,2,...,K
  \end{split},
\end{equation}
where ${D_k} \in {R^{{n_k} \times {M_k}}}$ is a dictionary with ${M_k}$ prototype atoms for the \emph{k}-th type of feature. ${X_k} \in {R^{{M_k} \times N}}$ and ${E_k} \in {R^{{n_k} \times N}}$ denote the SR coefficients and the reconstruction errors for the \emph{k}-th feature matrix ${Y_k}$, respectively. The joint error matrix \emph{E} is formed by concatenating the vertical columns of matrices ${E_1}$, ${E_2}$,...,${E_K}$.

As discussed in \cite{zhang2016robust,lang2012saliency}, the corresponding columns in the matrices ${E_1}$, ${E_2}$,...,${E_K}$ with the same index will be compelled to have similar magnitudes by imposing the $l_{2,1}$-norm minimization on the matrix \emph{E}. As for the RSR model, the optimization problem of MRSR can also be solved using LADMAP \cite{Lin2011Linearized,zhang2013learning}.

\subsection{Summary}

A close look at the aforementioned algorithms reveals that the essential difference among the SR models discussed above is where they apply the constraints, either on the representation coefficients, the reconstruction errors or on both. It can also be noticed that the traditional SR, NNSR, JSR and GSR models impose different constraints on the representation coefficients but the same least squared minimization constraint on the reconstruction errors. These SR models can thus be called \emph{least-squared-error-based} models. Differently, the RSR model replaces the conventional least-squares reconstruction error with a so-called sparse reconstruction error. Therefore, the RSR and MRSR models can be called \emph{sparse-error-based} models.

In contrast to those least-squared-error-based SR methods, using the sparse-error significantly improves the robustness of the RSR model against the non-Gaussian noise or sparse but strong corruptions, thereby facilitating practical applications. More importantly, many important features, including the detailed information contained in an image, can be denoted by the sparse error components obtained using the RSR model. Table. \ref{summary SPM} summarizes the previously mentioned sparse representation models.

Basically, the NNSR, JSR, GSR, RSR, and MRSR models somewhat improve the traditional SR model in various aspects, and they generally perform better than the SR model when applied to multi-sensor fusion applications. However, it is difficult to explain the suitability of a model for a specific application from the general point of view. Instead, we draw the conclusion based on the experimental results, which reveal that the RSR model seems to be more suitable for multi-focus image fusion; the NNSR and JSR are more suitable for multi-modality image fusion; and the GSR model can facilitate both as it achieves generally good results for these two applications. It is necessary to point out that the performance may be further improved if the dictionary of a model complies with the characteristics of the data. That is to say, it does not make sense to expect a universal dictionary that can enhance the performance of all the models. As a result, designing an appropriate dictionary for each model deserves further investigation.

\begin{table*}[]
	\centering
	\caption{Summary of the sparse representation models employed in multi-sensor image fusion.}
	\label{summary SPM}
	\begin{tabular}{cccc}
		\hline
		\multicolumn{2}{c}{Models}                        & Representation coefficients constrains                                                                   & Reconstruction error constrains                                                                             \\ \hline
		\multirow{4}{*}{Least-squared-error-based} & SR   & Sparisity constraint                                                                                     & \multirow{4}{*}{\begin{tabular}[c]{@{}c@{}}least squared minimization \\ contraint\end{tabular}}            \\ \cline{2-3}
		& NNSR & Sparisity and non-negativity contraint                                                                   &                                                                                                             \\ \cline{2-3}
		& JSR  & \begin{tabular}[c]{@{}c@{}}Sparisity common component and innovation\\ components contraint\end{tabular} &                                                                                                             \\ \cline{2-3}
		& GSR  & Group-sparisity contraint                                                                                &                                                                                                             \\ \hline
		\multirow{2}{*}{Sparse-error-based}        & RSR  & Sparisity contraint                                                                                      & Sparisity contraint                                                                                         \\ \cline{2-4} 
		& MRSR & Sparistiy contraint                                                                                      & \begin{tabular}[c]{@{}c@{}}Joint sparisity constraint cross error\\ matrices of multiple tasks\end{tabular} \\ \hline
	\end{tabular}
\end{table*}

\section{DICTIONARY LEARNING METHODS IN MULTI-SENSOR IMAGE FUSION}
Constructing a good dictionary is of fundamental importance for the performance of an SR-based image fusion method. Generally, there are two categories of methods to construct an over-complete dictionary. The first one uses some fixed basis \cite{yang2010multifocus,yang2012pixel-level}. In \cite{yang2010multifocus} for instance, an over-complete separable version of the DCT dictionary is constructed by sampling cosine waves with different frequencies. In \cite{yang2012pixel-level}, a hybrid dictionary consisting of a DCT basis, a wavelet `db1' basis, a Gabor basis and a ridgelet basis is constructed. Employing a fixed basis has the advantages of simplicity and fast implementation. Since this approach is not customized by using appropriate input image data, it may provide inferior performance for certain types of data and applications.

	The second category of methods is to construct an over-complete dictionary by using some learning methods, such as PCA, MOD and K-SVD \cite{Aharon2006}. These methods can be further divided into \emph{global-trained-dictionary-based} \cite{Liu2015A,liu2015simultaneous,yin2011multimodal,yang2012pixel-level} and \emph{adaptively-trained-dictionary-based} \cite{nejati2015multi-focus,wang2014fusion,zhang2016robust,zhang2013dictionary,yu2011image}, according to their employed training images. In the former methods, a public training database that generally contains many high-resolution images is employed to construct the training data for dictionary learning. For example, in \cite{Liu2015A}, the training data consists of 100,000 $8 \times 8$ patches, randomly sampled from a database of 40 high-quality images. While in the latter methods, the input images are directly used to construct the training data. For example, in \cite{wang2014fusion}, the training data for dictionary learning contains 20,000 $8 \times 8$
patches, which are randomly sampled from the source infrared and visible images. In \cite{nejati2015multi-focus}, local patches from the input multi-focus images are used as the training samples to learn a dictionary. In \cite{zhang2016robust}, the input image patches are directly employed to construct an over-complete dictionary. These dictionaries are adaptive to the input image data and thus have the potential to outperform the commonly used fixed dictionaries. Accordingly, these learned dictionaries are more widely adopted in SR-based image fusion. In the rest of this section, we review some dictionary learning methods used in multi-sensor image fusion\footnote{It should be noted that the methods to be discussed are adopted for the global-trained dictionaries as well as the adaptively-trained dictionaries.}.

\subsection{K-SVD based dictionary learning}
Let $Y = [{y_1},{y_2},...,{y_N}] \in {R^{n \times N}}$  be a training data matrix, where ${y_i} \in {R^n}$ is the \emph{i}-th sampled data vector. Our goal is to learn a dictionary $D = [{d_1},{d_2},...,{d_M}] \in {R^{n \times M}}$  and a sparse coefficient matrix $X = \left[ {{x_1},{x_2},...,{x_N}} \right] \in {R^{M \times N}}$, such that the product of \emph{D} and \emph{X} can approximate the original data matrix \emph{Y} efficiently. If \emph{X} were known, the over-complete dictionary \emph{D} could be obtained from the matrix \emph{Y} via solving
\begin{equation}\label{19}
  \mathop {\min }\limits_{D,X} \left\| {Y - DX} \right\|_F^2{\rm{        }}\ s.t.\ {\rm{  }}{\left\| {{x_i}} \right\|_0} \le \tau ,i = 1,2,...,N,
\end{equation}
where $\tau $ denotes the upper bound for the number of the non-zero entries in ${x_i}$ . The solution to (19) for both \emph{D} and \emph{X} can be obtained by using the popular dictionary learning algorithm K-SVD \cite{Aharon2006}, which iteratively alternates between two steps: sparse coding (find \emph{X}) and dictionary updating (find \emph{D}).

In the sparse coding step, \emph{D} is assumed to be fixed, and the optimization problem of (19) is reduced to a search for sparse representations with coefficients summarized in the matrix \emph{X}. For that, the criterion is rewritten as
\begin{equation}\label{20}
  \left\| {Y - DX} \right\|_F^2 = \sum\limits_{i = 1}^N {\left\| {{y_i} - D{x_i}} \right\|_2^2}.
\end{equation}

Therefore, the problem in (19) can be decoupled into \emph{N} optimization problems of the form
\begin{equation}\label{21}
  \mathop {\min }\limits_{{x_i}} \left\| {{y_i} - D{x_i}} \right\|_2^2{\rm{        }}\ s.t.\ {\rm{  }}{\left\| {{x_i}} \right\|_0} \le \tau ,i = 1,2,...,N.
\end{equation}
This problem can be efficiently solved by the MP \cite{mallat1993matching} and OMP \cite{bruckstein2009from} algorithms mentioned in Section II.

In the dictionary updating stage, the coefficient matrix \emph{X} and the dictionary \emph{D} are both assumed to be fixed. Only one column ${d_k}$  in the dictionary and the coefficients that correspond to it (i.e., the \emph{k}-th row of \emph{X}, denoted as $x_k^T$) are considered each time. For that, the multiplication $DX$ in (19) is decomposed into the sum of \emph{K} rank-1 matrices. During the updating, \emph{K}-1 terms are supposed to be fixed and one, i.e., the \emph{k}-th, remains in question. More specifically, the metric in (19) is rewritten as \cite{Aharon2006}
\begin{equation}\label{22}
  \begin{split}
  &\left\| {Y - DX} \right\|_F^2 = \left\| {Y - \sum\limits_{j = 1}^M {{d_j}x_j^T} } \right\|_F^2\\
  &= \left\| {\left( {Y - \sum\limits_{j \ne k} {{d_j}x_j^T} } \right) - {d_k}x_k^T} \right\|_F^2 = \left\| {{E_k} - {d_k}x_k^T} \right\|_F^2
  \end{split},
\end{equation}
where ${E_k}$ stands for the error for all the \emph{N} samples when the \emph{k}-th atom is removed. Minimizing the function in (22) is equivalent to finding a rank-1 matrix that closely approximates the error term ${E_k}$ in Frobenius norm. The rank-1 matrix is described by the atom ${d_k}$ and the row vector $x_k^T$. These can be obtained simply by using singular value decomposition (SVD) on ${E_k}$. Moreover, to ensure the sparsity of the vector $x_k^T$, some modifications are further performed on (22). More details can be found in \cite{Aharon2006}.

\subsection{MOD-like based Dictionary learning}
In \cite{zhang2013dictionary}, the authors present a dictionary learning method (termed as \emph{MODJSR}) for the JSR model. Similar to the traditional dictionary learning methods using K-SVD, MODJSR is also implemented by alternating the sparse coding stage and the dictionary updating stage. In the second stage, dictionary updating is performed as a problem by the \lq\lq Landweber\rq\rq\ update \cite{Landweber1951An} with an initial point obtained by the method of optimal directions (MOD). This method is shown to have higher computational efficiency than the K-SVD method.

Suppose ${Y_k} \in {R^{n \times L}}(k = 1,...,K)$ are signals from the same ensemble, i.e., from different source images of the same scene. Motivated by dictionary learning for the standard SR, the dictionary learning method, MODJSR, for the JSR model is defined as \cite{zhang2013dictionary}
\begin{equation}\label{23}
  \mathop {\min }\limits_{D,X} \frac{1}{2}\left\| {Y - \underline{D}X} \right\|_F^2{\rm{      }}\ s.t.\ {\rm{    }}{\left\| {{x_t}} \right\|_0} \le \tau ,t = 1,2,...,L.
\end{equation}
Here, the data set matrix $Y \in {R^{nK \times L}}$, the dictionary matrix $ \underline{D}\in {R^{nK \times (K + 1)M}}$ and the coefficient matrix $X \in {R^{(K + 1)M \times L}}$ are constructed as in (7), (8), and (9), respectively. $\tau $ denotes the maximal number of non-zeros coefficients used in each column of \emph{X}.

Adopting the block-coordinate descent idea, an alternating strategy is used to solve (23) with two stages. The first stage employs a joint sparse coding. That is, fixing the dictionary \emph{D}, the joint sparse coefficient matrix \emph{X} can be obtained by solving (10) via OMP \cite{bruckstein2009from} to take advantage of its simplicity and fast execution.

The second stage updates the dictionary. Fixing the joint sparse coefficient matrix \emph{X}, the dictionary $\underline D$  in (23) could be updated simply by $\underline{\hat D}  = Y{X^T}{(X{X^T})^{ - 1}}$ with MOD. However, $X{X^T}$ may not always be full rank. The majorization method could be also directly employed, but it is slow due to using the \lq\lq Landweber\rq\rq\  update which is a gradient update. If the dictionary is updated by the \lq\lq Landweber\rq\rq\ update, the initial point can be obtained by MOD. Then \emph{D} is found by solving \cite{zhang2013dictionary}
\begin{equation}\label{24}
  \begin{split}
  \mathop {\min }\limits_D f(D) &= \mathop {\min }\limits_D \frac{1}{2}\left\| {Y - \underline{D}X} \right\|_F^2\\
  &= \mathop {\min }\limits_D \sum\limits_{k = 1}^K {\frac{1}{2}\left\| {{Y_k} - D\left( {{X^C} + X_k^U} \right)} \right\|_F^2}\\
  \end{split}.
\end{equation}

The optimum of the objective function satisfies
\begin{equation}\label{25}
  0 = \frac{d}{{dD}}f(D).
\end{equation}

Hence,
\begin{equation}\label{26}
  W = DH,
\end{equation}
where $W = \sum\nolimits_{k = 1}^K {{Y_k}{{\left( {{X^C} + X_k^U} \right)}^T}} $ and $H = \sum\nolimits_{k = 1}^K {\left( {{X^C} + X_k^U} \right)} {\left( {{X^C} + X_k^U} \right)^T}$. Since \emph{X} is sparse, the non-zero elements of \emph{H }are often concentrated on the diagonal and ${H_{ii}} \ge 0$ $(i = 1,...,M)$, $rank(H) = M$ holds with high probability \cite{HornJ85} due to Diagonal Dominance theory. When $rank(H) = M$, the dictionary \emph{D} is simply updated by $D = W{H^{ - 1}}$. Otherwise, it is updated by the \lq\lq Landweber\rq\rq\ rule as \cite{zhang2013dictionary,Landweber1951An}
\begin{equation}\label{27}
  {D^{[k + 1]}} = {D^{[k]}} + \frac{1}{\sigma }\left( {W - {D^{[k]}}H} \right){H^T},
\end{equation}
where $\sigma $ is a constant satisfying $\sigma  > {\left\| {{H^T}H} \right\|_F}$. A good initial point, obtained by MOD and given by ${D^{[0]}} = W{H^\diamondsuit }$ is employed while updating the dictionary updating for higher computation efficiency. Here ${H^\diamondsuit }$ is computed as ${H^\diamondsuit } = U{\Sigma ^\dag }{U^T}$ and the matrices \emph{U} and $\Sigma $ result from the SVD of the matrix \emph{H}, i.e., $H = U\Sigma {U^T}$.
\subsection{PCA and joint patch clustering based dictionary learning}
Since the connection of sparsity and clustering was shown to be desirable in image restoration tasks \cite{kim2016joint,mairal2009non-local}, some new dictionary learning frameworks combined with clustering of non-local patches were recently presented \cite{Chatterjee2009Clustering,dong2011sparsity-based}. Motivated by clustering-based dictionary learning techniques, the authors presented an efficient dictionary learning method based on a joint patch clustering for multi-modal image fusion in \cite{kim2016joint}. This is also the first attempt towards applying clustering-based dictionary learning to image fusion.

	Conventional dictionary learning methods based on K-SVD, such as the ones discussed in the previous subsections, generally produce redundant or highly structured dictionaries \cite{kim2016joint}. The proposed dictionary learning in \cite{kim2016joint} aims to remove the redundancy while maintaining or improving the quality of the multimodal image fusion. Under an assumption that common image structures are distributed across the source images from different sensor modalities, patches from different source images are clustered together according to local structural similarities. Then sub-dictionaries that best describe the underlying structure of each cluster by using only a few principal components are constructed. Finally, these sub-dictionaries are combined to form a final dictionary.

	Since each sub-dictionary consists only of a few principal components of each joint patch cluster, the final dictionary constructed ends up with much smaller size than those learned by K-SVD. Although it is more compact, the constructed dictionary still contains the most informative components from each joint patch cluster. As a result, the computational complexity of the subsequent fusion method is greatly reduced while the fusion performance is maintained.
\subsection{Dictionary learning for adaptive sparse representation}
\begin{figure}[h]
  \centering
  \includegraphics[width=0.95\linewidth]{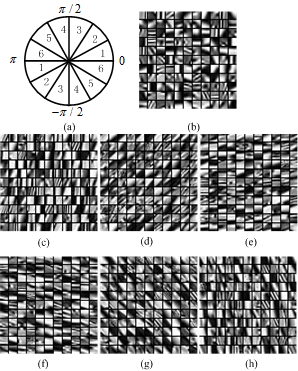}\\
  \caption{Learning sub-dictionaries in the ASR model. (a) Illustration of the dominant orientation division; (b)-(h) Learned sub-dictionaries $\left\{ {{D_k}|k = 0,1,...,6} \right\}$, respectively. (Credit to [22])}\label{7}
\end{figure}

\begin{figure*}[!t]
  \centering
  \includegraphics[width=0.95\linewidth]{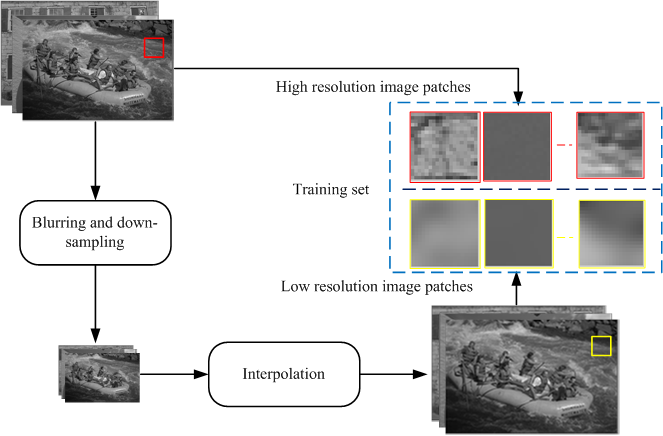}\\
  \caption{Procedure to construct the training sets for the coupled dictionaries.}\label{8}
\end{figure*}

	In the traditional SR models introduced in Section II, a highly redundant dictionary is always needed to satisfy signal reconstruction requirements since the structures vary significantly across different image patches. However, this may result in potential visual artifacts as well as high computational cost. To address this problem, the authors in \cite{liu2015simultaneous} introduced an adaptive sparse representation (ASR) model, in which a set of more compact sub-dictionaries are learned from numerous high-quality image patches. These patches have already been pre-classified into several corresponding categories based on their gradient information.

Let $P = \left\{ {{p_1},{p_2},...,{p_N}} \right\}$ be a training data matrix, where ${p_i} \in {R^n}$ is the \emph{i}-th sampled data or image patch. The patches in set \emph{P} are first classified into \emph{K} categories $\left\{ {{P_k}|k = 1,2,...,K} \right\}$ according to their dominant gradient directions. Then a total of $K + 1$ sub-dictionaries $\left\{ {{D_o},{D_1},...,{D_K}} \right\}$ are obtained, in which ${D_0}$ is learned from all the patches in \emph{P} having no clear dominant directions, whereas $\left\{ {{D_k}|k = 1,2,...,K} \right\}$ is learned from the patches in each corresponding subset $\left\{ {{P_k}|k = 1,2,...,K} \right\}$ that have specific dominant directions described by category \emph{k}. In this method, the dominant gradient direction of each signal ${y_i}$ is first computed, after which the sub-dictionary ${D_{{k_i}}}$ is adaptively selected as the dictionary. An example of the ASR dictionary learning with $K = 6$ is shown in Fig. 7.

\begin{table*}[]
	\centering
	\caption{Comparison of different dictionary learning methods}
	\label{dictionary learning}
	\begin{tabular}{ccccc}
		\hline
		& number of dictionaries                                 & redundancy & applied model       & computation efficiency \\ \hline
		K-SVD-DL   & 1                                                      & high       & SR, RSR, MRSR       & low                    \\ \hline
		MOD-DL     & 1                                                      & high       & JSR                 & high                   \\ \hline
		PCA-DL     & 1 (multiple sub-dictionaries)                                   & low        & SR, GSR, RSR, MRSR  & high                   \\ \hline
		ASR-DL     & $>1$ (specific dominant directions) +  1 (common) & low        & SR, RSR, MRSR       & medium                 \\ \hline
		Coupled-DL & 2                                                      & high       & SR, NNSR, RSR, MRSR & low                    \\ \hline
	\end{tabular}
\end{table*}

\subsection{Coupled dictionaries learning}
In \cite{yang2010image}, sparse representation was applied to single image super-resolution. The main idea of the method is to assume that the upsampled low-resolution (LR) and high-resolution (HR) image patch pairs share the same sparse coefficients with respect to their own dictionaries. Recently, this idea was applied to multi-sensor image fusion \cite{yin2013simultaneous,iqbal2012unification,ren2015super-resolution} as well as pan-sharpening\cite{guo2014an,zhu2013a}.

In order to construct a pair of coupled dictionaries, two training sets for the LR and HR dictionaries are first constructed from the same set of HR training images\footnote{For pan-sharpening, the training sets may be constructed from the HR panchromatic source images.} as shown in Fig. 8 and explained thereafter. Each high-resolution image \emph{I} is blurred and down-sampled (with a user-defined factor) to generate a LR image. The latter is then up-sampled back to the original size using Bicubic interpolation and the resulting image is seen as a LR image. A pair of training sets $\left\{ {y_i^H \in {R^n}|i = 1,2,...,N} \right\}$, $\left\{ {y_i^L \in {R^n}|i = 1,2,...,N} \right\}$ are thus created by extracting patches from the original HR image \emph{I} and its degraded LR version, respectively, in which $y_i^H$ and $y_i^L$ with the same index \emph{i} correspond to the same spatial position in the HR and LR images. Based on the assumption that sparse coefficients of the LR image patch $y_i^L$ over the LR dictionary ${D_L} \in {R^{n \times M}}$ are the same as those of the HR image patch $y_i^H$ over the HR dictionary ${D_H} \in {R^{n \times M}}$, the coupled dictionaries ${D_H}$ and ${D_L}$ can be learned by solving the following optimization problem \cite{yin2013simultaneous}
\begin{equation}\label{28}
  \begin{split}
  &\left\{ {{D_H},{D_L},X} \right\}= \\
  &\mathop {\arg \min }\limits_{{D_H},{D_L},X} \sum\limits_{i = 1}^N {\left\| {y_i^H - {D_H}{x_i}} \right\|_2^2}+ \sum\limits_{i = 1}^N {\left\| {y_i^L - {D_L}{x_i}} \right\|_2^2} \\
  &s.t.\ {\rm{   }}\forall i{\rm{  }}{\left\| {{x_i}} \right\|_0} \le \tau \\
  \end{split},
\end{equation}
where $X = \left[ {{x_1},{x_2},....,{x_N}} \right] \in {R^{M \times N}}$ is the matrix containing the sparse coefficients, and $\tau $ controls the sparsity level. By introducing auxiliary variables ${Y^H} = [y_1^H,y_2^H,...,y_K^H] \in {R^{n \times N}}$, ${Y^L} = [y_1^L,y_2^L,...,y_N^L] \in {R^{n \times N}}$, $Y = {\left[ {{{\left( {{Y^H}} \right)}^T},{{\left( {{Y^L}} \right)}^T}} \right]^T} \in {R^{2n \times N}}$, and $D = {\left[ {{{\left( {{D_H}} \right)}^T},{{\left( {{D_L}} \right)}^T}} \right]^T} \in {R^{2n \times M}}$, problem (28) is equivalently transformed to (19) and can thus be efficiently solved by K-SVD.

\subsection{Summary}

As discussed in this section, many dictionary learning methods have been presented or applied to multi-sensor image fusion. Among these methods, the K-SVD method, thanks to its simplicity and generalization, is the most broadly adopted by the existing SR-based fusion methods. To some extent, the learning procedure of the ASR dictionary and the coupled dictionary are also K-SVD like based on the same principle. It is worthwhile pointing out that each dictionary learning method has its pros and cons, meaning that there is no universal dictionary that suits all applications. 

Using these methods, a globally-trained dictionary or an adaptively-trained dictionary can be generated during the fusion process. These learned dictionaries are adaptive to the input image data and usually perform better than the fixed dictionaries in terms of the extraction and representation of significant features in an image. However, these learned dictionaries generally contain a large number of atoms in order to accurately reconstruct an input image patch. This increases the redundancy among the dictionary atoms and thus degrades the subsequent fusion performance to some extent. Moreover, this also increases the computational complexity of a fusion method. In Table. 3, we compare some existing dictionary learning methods with respect to the number of sub-dictionaries, redundancy, applicable model and consumed computation power. Nevertheless, how to learn a dictionary with a fixed small number of atoms and yet maintain a good representation capability for different SR models and fusion applications is desirable and still a challenging problem in multi-sensor image fusion.

\section{APPLICATIONS OF DIFFERENT SR-BASED FUSION METHODS}
So far, SR-based image fusion methods have been used in a wide variety of applications, such as multi-focus image fusion, and multi-modality (e.g., infrared and visible light) image fusion. These applications are targeting different fusion goals, and thus have different fusion strategies. In this section, we will review some applications of SR-based fusion methods for fusing multi-focus images or infrared with visible images.

\begin{figure*}[!t]
  \centering
  \includegraphics[width=0.95\linewidth]{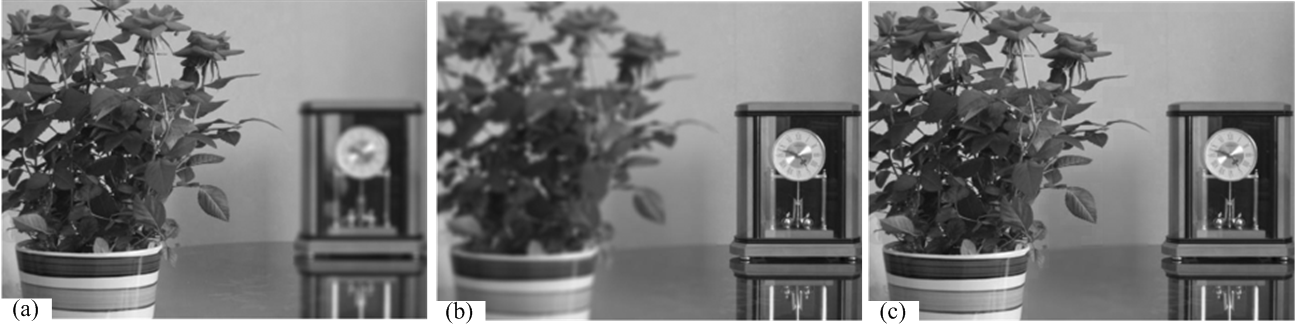}\\
  \caption{Illustration of multi-focus image fusion. (a) Focus on the flower; (b) Focus on the clock; (c) Fused image with full-focus.}\label{9}
\end{figure*}

\subsection{SR-based multi-focus image fusion}
Due to the limited depth-of-focus of optical lenses in CCD devices, it is often not possible to obtain an image that contains all of the relevant objects in focus. As shown in Fig. 9 \footnote{The test images in Fig. 9 and the soon Fig. 11 are downloaded from http://home.ustc.edu.cn/$~$liuyu1}, this issue can be overcome by multi-focus image fusion, in which several images with different focus points (e.g., Fig. 9(a) and Fig. 9(b)) are combined to form a composite image (e.g., Fig. 9(c)) with full-focus. The basic requirement for multi-focus image fusion is that only the focused regions should be extracted from the given multi-focus input images and then preserved in the fused image, while all of the defocused regions should be discarded.

As shown in Fig. 2 in Section 1, the SR-based multi-focus image fusion generally involves the following steps: (1) Divide the source images into a larger number of image patches of the same size (e.g., $8 \times 8$). In order to reduce block artifacts and improve robustness to mis-registration, a sliding window at a step length of a fixed number of pixels (e.g., one pixel) is also often used in this step. That is to say, these patches overlap by a fixed number of pixels along the horizontal and vertical directions, respectively. (2) Re-order each of these patches as a vector of \emph{n}-dimensions (e.g., $n = 8 \times 8 = 64$). (3) Sparsely code these vectors via different SR models and pre-constructed dictionaries introduced in Sections II and III. The traditional SR model introduced in Section II.A is the most widely used in multi-focus image fusion. The dictionaries directly learned from a set of training images with high-resolution using K-SVD are also the most popular in these methods. (4) Define activity levels and then construct the fused image with different fusion rules.

Activity level reflects the importance of each local image patch. Particularly, for multi-focus image fusion, the activity level should reflect the focus information of each image patch. In SR-based multi-focus image fusion methods, the activity level is generally defined as the $l_0$-norm, $l_1$-norm or the $l_2$-norm of the sparse coefficient vector for each image patch, i.e.,
\begin{equation}\label{29}
A({p_{{k_i}}}) = {\left\| {{x_{{k_i}}}} \right\|_j}
\end{equation}
where ${p_{{k_i}}}$ denotes the \emph{i}-th patch from the \emph{k}-th source image, ${x_{{k_i}}}$ denotes the representation coefficient vector corresponding to the patch ${p_{{k_i}}}$, and $j=0$, 1, or 2 describes which norm function is employed to define the activity level.

Sometimes, relatively more sophisticated activity levels are also defined. For example, in \cite{nejati2015multi-focus}, the correlation between the sparse representation of the input images and the pooled features obtained in the previous dictionary learning phase is used as the decision map for the fusion. As opposed to most SR-based multi-focus image fusion methods employing the sparse representation coefficients to define activity level, the fusion method presented in \cite{zhang2016robust} employs the sparse reconstruction error, more specifically, the $l_2$-norm of each column vector in the sparse error matrix obtained by the RSR model, to define the activity level for each source image patch.

There are two different ways to construct the fused image after the activity level of each image patch is determined. Accordingly, different SR-based multi-focus image fusion methods are divided into two categories, \emph{transform-domain-based} and \emph{spatial-domain-based}. In the transform-domain-based fusion methods \cite{yang2010multifocus,liu2015simultaneous,li2012group-sparse,yang2012pixel-level,ibrahim2015pixel,Zhang2014A,yao2012omp,yao2013image,Yang2016Simultaneous}, the representation coefficients of fused image patches are first obtained from the corresponding representation coefficients of source image patches according to their activity levels. Then the fused image patches are constructed by multiplying the pre-defined dictionary with the obtained representation coefficients. On the other hand, in the spatial-domain-based fusion methods \cite{nejati2015multi-focus,zhang2016robust}, the fused image patches are directly extracted from the source image patches according to their activity levels.

	In general, both the maximum-selection and weighted-averaging fusion rules (or fusion strategies) might be employed to determine the fused image patches or their representation coefficients. However, in the SR-based multi-focus fusion methods, the maximum-selection fusion rule is more popular. In this approach, the fused image patch or its sparse representation is generally selected from the input image patch or its sparse representation with the highest activity level. Some state-of-art SR-based multi-focus image fusion methods are summarized in Table 4.

\begin{table*}[!t]
  \centering
  \caption{Some state-of-the-art SR-based multi-focus image fusion methods.}\label{2}
   \begin{tabular}{c c c c c}
   \toprule
   \multicolumn{2}{c}{Method} & \multicolumn{1}{c}{Model} & \multicolumn{1}{c}{Dictionary} &\multicolumn{1}{c}{Fusion rule}\\
   \midrule
    & \makecell{\cite{yang2010multifocus,yang2012pixel-level},\\\cite{ibrahim2015pixel,Zhang2014A}}& \makecell{SR} &\makecell{Learned from a set of images \\\cite{yang2010multifocus,yang2012pixel-level,Zhang2014A}\\Fixed DCT basis \cite{yang2010multifocus,yang2012pixel-level,ibrahim2015pixel}\\Fixed hybrid basis\cite{yang2012pixel-level}\\Fixed hybrid basis\cite{yang2012pixel-level}}&\makecell{Maximum $l_1$-norm selection of representation \\coefficient vectors \cite{yang2010multifocus}\\Maximum selection of absolute coefficient\\vector entries \cite{yang2012pixel-level}\\Maximum $l_2$-norm selection of representation\\coefficient vectors \cite{ibrahim2015pixel}\\Weighed averaging of representation coefficient\\vectors \cite{Zhang2014A}}\\
   \cline{2-5}\\
   \makecell{Transform-domain-based}& \makecell{\cite{li2012group-sparse}}&\makecell{Group SR}&\makecell{Learned from a set of images}&\makecell{Maximum $l_2$-norm selection of representation\\coefficient vectors}\\
   \cline{2-5}\\
   &\makecell{\cite{liu2015simultaneous}}&\makecell{Adaptive SR}&\makecell{Multiple dictionaries with different\\dominant directions learned from\\a set of images}&\makecell{Maximum $l_1$-norm selection of representation\\coefficient vectors}\\
   \cline{2-5}\\
   &\makecell{\cite{yao2012omp,yao2013image}}&\makecell{JSR}&\makecell{Learned from source images}&\makecell{Summing of representation coefficient vectors}\\
   \cline{2-5}\\
   &\makecell{\cite{Yang2016Simultaneous}}&\makecell{Extended JSR}&\makecell{Learned from a set of images}&\makecell{Maximum $l_1$-norm selection of representation \\coefficient vectors}\\
   \hline
   &\makecell{\cite{nejati2015multi-focus}}&\makecell{SR}&\makecell{Learned from source images}&\makecell{Maximum correlation between the sparse\\ representations of input source images and the\\ training pooled features}\\
   \cline{2-5}\\
   \makecell{Spatial-domain-based}&\makecell{\cite{zhang2016robust}}&\makecell{RSR}&\makecell{Data itself}&\makecell{Maximum $l_2$-norm selection of spare \\reconstruction error vectors}\\
   \cline{2-5}\\
   &\makecell{\cite{zhang2016robust}}&\makecell{Multi-task\\RSR}&\makecell{Data itself}&\makecell{Maximum $l_2$-norm selection of joint sparse \\reconstruction error vectors}\\
   \bottomrule
   \hline
   \end{tabular}

\end{table*}
\subsection{SR-based multi-modality image fusion}
It is becoming more common to employ multiple types of imaging sensors in video surveillance to improve the robustness, in which visible light and infrared imaging sensors are normally combined. Image fusion allows the information captured by these different sensors to be sufficiently and effectively integrated to create a composite image, containing more useful information than any of the individual input images. This image can be used to better interpret the scene \cite{zhang2015multisensor}. Multi-modality image fusion has also been widely applied to many other fields such as medical imaging.

	A video surveillance application is shown in Fig. 10 (a), where the moving person is evident in the image taken by the infrared video camera. However, the scene environment (e.g., the hedges and the shrubs) is better displayed in the visible-light image (Fig. 10(b)), in which the moving targets are difficult to see. By fusing the two input images, the moving target from the infrared camera and the background scene (or the environment) from the visible light camera are well integrated. As shown in Fig. 10(c), the fused image clearly shows that there is a man in the scene.
\begin{figure*}[!t]
  \centering
  \includegraphics[width=0.95\linewidth]{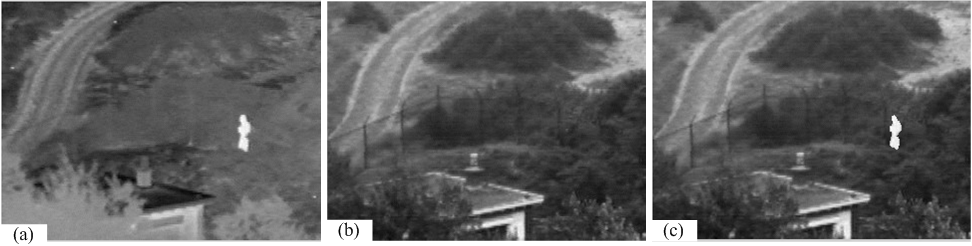}\\
  \caption{Illustration of infrared and visible image fusion. (a) Infrared image; (b) Visible light image; (c) Fused image.}\label{10}
\end{figure*}
	SR has also been applied to multi-modality image fusion, including infrared and visible light  sensors \cite{Liu2015A,wang2014fusion,zhang2013dictionary,kim2016joint,yin2011multimodal,yu2011image,lu2014the}. Due to different imaging technologies of the sensors, these multi-modality images of the same scene captured by different image sensors provide redundant and complementary information. The basic job of a multi-modality image fusion approach is to properly employ the redundant and complementary information available from the different input images \cite{zhang2011similarity-based}.

	Interestingly, this notion maps well into the JSR model and this is reflected by the fact that, in addition to the traditional SR model, the JSR model is popular in multi-modality image fusion \cite{zhang2013dictionary,yin2011multimodal,yu2011image,Yang2014Visual}. The reason for this is that in the JSR model, all the signals from the same ensemble are automatically decomposed into a common component that is shared by all the signals and an innovation component that describes each individual signal. The common component describes the redundant information among all the signals, while the innovation component describes the complementary information \cite{yu2011image}. Accordingly, JSR already extracts the required information needed for fusion. In the subsequent fusion phase, the innovation components for the input images are combined together by using a weighted-averaging \cite{zhang2013dictionary,yu2011image} or a summing \cite{yin2011multimodal,Yang2014Visual} fusion strategy. The final fused image is obtained by integrating the common component shared by all the input images into the previously combined innovation component.
\begin{table*}[!t]
  \centering
  \caption{Some state-of-the-art SR-based multi-modality image fusion methods.}\label{3}
  \begin{tabular}{c c c c}
   \toprule
   Methods&Model&Dictionary&Fusion rule\\
   \midrule
   \makecell{\cite{Liu2015A,kim2016joint,lu2014the},\\\cite{ding2013research,yin2015sparse,Liu2014Medical}}&\makecell{SR}&\makecell{Learned from a set of images\\{\cite{Liu2015A,yin2015sparse,Liu2014Medical}}\\Learned from source images\\{\cite{kim2016joint,lu2014the,ding2013research}}}&\makecell{Maximum $l_1$-norm selection of \\representation coefficient vectors \cite{Liu2015A,Liu2014Medical}\\Maximum $l_2$-norm selection of \\representation coefficient vectors \cite{yin2015sparse}\\Maximum selection of (absolute) \\coefficient vector entries \cite{lu2014the,ding2013research}\\Summing of representation coefficient vectors \cite{kim2016joint}}\\
   \hline
   \makecell{\cite{li2012group-sparse}}&\makecell{Group SR}&\makecell{Learned from a set of images }&\makecell{Maximum $l_2$-norm selection of \\representation coefficient vectors}\\
   \hline
    \makecell{\cite{liu2015simultaneous}}&\makecell{Adaptive SR}&\makecell{Multiple dictionaries with different \\dominant directions learned from \\a set of images }&\makecell{Maximum $l_1$-norm selection of \\representation coefficient vectors}\\
    \hline
        \makecell{\cite{wang2014fusion}}&\makecell{NNSR}&\makecell{Learned from source images }&\makecell{Maximum $l_1$-norm \&\ sparseness \\selection of representation coefficient vectors}\\
    \hline
        \makecell{\cite{zhang2013dictionary,yin2011multimodal},\\\cite{yu2011image,Yang2014Visual}}&\makecell{JSR}&\makecell{Learned from a set of images \cite{yin2011multimodal,Yang2014Visual}\\Learned from source images \cite{zhang2013dictionary,yu2011image}}&\makecell{Summing of representation coefficient vectors \cite{yin2011multimodal,Yang2014Visual}\\Weighted averaging of representation coefficient \\vectors \cite{zhang2013dictionary,yu2011image}}\\
    \bottomrule
  \end{tabular}
\end{table*}
\begin{figure*}[!t]
  \centering
  \includegraphics[width=0.95\linewidth]{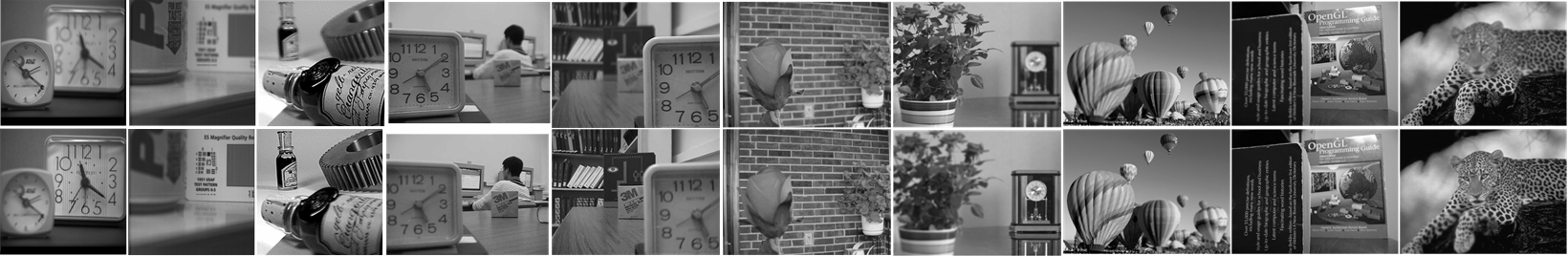}\\
  \caption{10 pairs of multi-focus test images. The top row contains 10 input images with the focus on the left part, and the bottom row contains the corresponding input images with the focus on the right part.}\label{11}
\end{figure*}
\begin{figure*}[!t]
  \centering
  \includegraphics[width=0.95\linewidth]{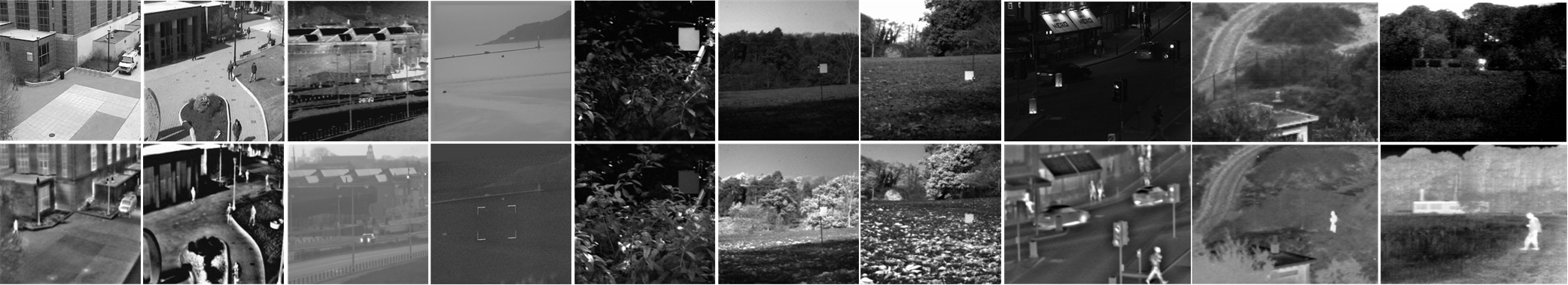}\\
  \caption{10 pairs of multi-modality test images. The top row contains 10 visible input images, and the bottom row contains the corresponding infrared input images.}\label{12}
\end{figure*}
	Finally, it should be noted that almost all SR-based multi-modality image fusion methods are transform-domain-based. This may result from the fact that patches from the multi-modality input images corresponding to the same spatial positions have greatly diverse characters because of the different sensor technologies. Subsequently, many spatial artifacts will be introduced during the fusion if a spatial-domain-based method is adopted which tends to produce higher activity levels. Alternatively, a transform-domain-based method may reduce the artifacts to some extent. Table 5 summarizes some state-of-art SR-based multi-modality image fusion methods.
\section{EXPERIMENTS AND ANALYSIS}
As discussed in the previous sections, SR models, learned dictionaries and activity levels are three important issues in SR-based fusion methods. In this section, we will discuss the impacts of these three components on the fusion performance in the context of the previous two applications. For this purpose, we employ two sets of test images, as shown in Fig. 11 and Fig. 12. The two sets of test images contain 10 pairs of multi-focus images and 10 pairs of infrared and visible images, respectively. We employ the mutual information (\emph{MI}) \cite{qu2002information}, the gradient preservation quality metric $Q_G$ \cite{xydeas2000objective}, the structure similarity (SSIM) fusion quality metric $Q_S$, and two phase-congruency fusion quality metrics $Q_{ZP}$
 \cite{liu2008a} and $Q_{PC}$ \cite{Zhang2013Multimodality} to evaluate different fusion methods.

	In these experiments, the SR-based fusion methods are applied on a patch by patch basis. That is, the source images are first divided into many patches of the same size and then these patches are fused. The size of the patches is set to $8 \times 8$ as referring to the experimental results in \cite{yang2010multifocus}. Accordingly, the size of the dictionary atoms is also set to $8 \times 8$. In addition, in order to improve the robustness to mis-registration and reduce the spatial artifacts, a sliding window technology is employed, i.e., the patches overlap by one pixel.
\begin{table*}[!t]
  \centering
  \caption{Different fusion methods with various SR models and their key parameters.}\label{4}
  \begin{tabular}{c | c | c}
   \toprule
   Model&Dictionary&Fusion rule\\
   \midrule
   SR \cite{wright2009robust,yang2010multifocus,Liu2015A} &Learned from a set of images using K-SVD method \cite{Liu2015A}&\\
   \cline{1-2}
   ASR \cite{liu2015simultaneous}&\makecell{Multiple dictionaries \cite{liu2015simultaneous} with different dominant directions\\ learned from a set of images} &\\
   \cline{1-2}
   GSR \cite{li2012group-sparse}&\makecell{Learned from a set of images using the patch-clustering-based\\ method \cite{kim2016joint}} &\makecell{Maximum $l_1$-norm selection of \\representation coefficient vectors}\\
   \cline{1-2}
   NNSR \cite{wang2014fusion}&\makecell{Learned from a set of images using the method in \cite{dong2016hyperspectral}} &\\
   \cline{1-2}
      JSR \cite{zhang2013dictionary,Baron2012Distributed,yu2011image}&\makecell{Learned from a set of images using K-SVD method \cite{Liu2015A}} &\\
   \hline
      RSR [28]&\makecell{Learned from a set of images using K-SVD method \cite{Liu2015A}} &\makecell{Maximum $l_2$-norm of sparse \\reconstruction errors}\\
   \bottomrule
  \end{tabular}
\end{table*}

\subsection{SR models}
Next, the impact of different sparse representation models (listed in Table 6\footnote{The dictionaries used in the models mentioned in Table 6 are learned from a database containing 24 high-resolution training images that are downloaded from http://r0k.us/graphics/kodak/.}) on the fusion performance will be discussed. Table 7 provides the scores of the different fusion methods on the two sets of test images, it indicates that the sparse representation model has a great effect on the fusion performance. As shown in Table 7, fusion performance varies significantly with the employed sparse representation model in an image fusion method. It also shows that the GSR performs the best among the six models considered here. In terms of most quality metrics, it achieves the highest scores for the fusion of multi-focus images as well as for the fusion of infrared and visible images. This may be due to the cluster structure sparsity prior employed in the GSR model. In addition to GSR, RSR and NNSR could also achieve satisfactory results when applied to multi-focus image fusion and multi-modality image fusion, respectively. However, for multi-modality image fusion, JSR could not achieve a satisfactory result as it did in \cite{zhang2013dictionary}. This might be due to the employed dictionary KSVD-512 that was learned for SR rather than for JSR.
\begin{table}[h]
  \centering
  \caption{Performance of different SR models on the two sets of test images. Scores for all image pairs in each dataset are averaged.}\label{5}
  \begin{tabular}{c | c | c c c c c}
   \toprule
   \makecell{Test \\images}&Models&\emph{MI} & $Q_G$ &$Q_S$ &$Q_{ZP}$ & $Q_{PC}$\\
   \midrule
   \multirow{6}{*}{\makecell{Multi-\\focus \\images}}& SR & 4.1267 & 0.7584 & 0.5008 & 0.9533 & 0.6846\\
   & ASR & 4.0889 & 0.7548 & 0.4976 & 0.9444 & 0.6773\\
   & GSR & 4.6534 & \textbf{0.7696} & \textbf{0.5097}	& 0.9587 & \textbf{0.6940}\\
   & NNSR & 4.0504 & 0.7565 & 0.4994 & 0.9574 & 0.6615\\
   & JSR & 4.6081 & 0.7666 & 0.5108 & 0.9565 & 0.6934\\
   & RSR & \textbf{4.8720} & 0.7691 & 0.5024	& \textbf{0.9681} &0.6916\\
   \hline
   \multirow{6}{*}{\makecell{Visible-\\infrared \\images}}& SR & 2.3239 & 0.6192	& 0.4225 & 0.8340 & 0.4208\\
   & ASR & 2.2411	& 0.5966	& 0.4154	& 0.8284	& 0.4112\\
   & GSR & \textbf{3.0451}	& \textbf{0.6346}	& 0.4240	& 0.8658	& 0.4486\\
   & NNSR & 2.8963	&0.6194	&0.4152	&\textbf{0.9025}	&\textbf{0.4699}\\
   & JSR & 2.4258	& 0.6178	& 0.4205	& 0.7815	& 0.3992\\
   & RSR & 2.7403	& 0.6335	& \textbf{0.4320}	& 0.7936	& 0.4129\\
  \bottomrule
  \end{tabular}
\end{table}
\subsection{Dictionary construction}
In this part, we will study the effect of the employed dictionary on the fusion performance. In all the experiments conducted, we employ the traditional SR model, and the maximum l1-norm as the fusion rule during the fusion process. Moreover, we test two kinds of over-complete dictionaries on the two sets of test images. The first is a 2-D over-complete DCT dictionary of size 512 (DCT-512, for short) \cite{yang2010multifocus}. The second includes four global trained dictionaries of size 128, 256, 512, and 1024. The four dictionaries (KSVD-128, KSVD-256, KSVD-512, and KSVD-1024, for short) are all learned from image samples using the iterative K-SVD algorithm \cite{Aharon2006}. The training data consists of 50,000 $8 \times 8$  patches, randomly taken from the database mentioned in the previous Section V.A. We also test three sets of adaptively trained dictionaries (denoted by D$_{vi}$-512,D$_{ir}$-512, and D$_{joint}$-512) on the infrared-visible test image set (i.e., the second set of test images). Each dictionary in the D$_{vi}$-512 set consists of 512 atoms and is learned from the corresponding visible input image in the second set of test images by using the iterative K-SVD algorithm. Similarly, each dictionary in the D$_{ir}$-512 set is learned from the corresponding infrared input image, and each dictionary in the D$_{joint}$-512 set is learned from the corresponding visible and infrared test images.

Table 8 provides the fusion scores of different dictionaries on the two sets of test images. According to Table 8: (1) As expected, the global learned dictionaries usually perform better than the fixed DCT dictionary. (2) The adaptively trained dictionaries in D$_{vi}$-512 and D$_{joint}$-512 sets, especially the ones dictionaries in the former set, perform competitively with the global dictionary having the same number of atoms when applied to multi-modality image fusion. However, the dictionaries in the D$_{ir}$-512 set that are adaptively learned from the infrared input images do not perform better than the global learned dictionary and the ones in D$_{vi}$-512 and D$_{joint}$-512 sets. This may be due to the fact that fewer patches in the infrared images contain significant structures. As a result, the dictionaries in the D$_{ir}$-512 set have weak representation power and reduce the fusion performance. In contrast, the visible input images contain many more patches with significant structures. Correspondingly, the dictionaries in the D$_{vi}$-512 set seem to achieve better fusion performance. (3) It can also be argued that the number of dictionary atoms have a great impact on the fusion performance. As shown in Table 8, the dictionary KSVD-512 obtains the highest fusion performance among the four global dictionaries studied when applied to multi-focus image fusion as well as multi-modality image fusion. For the dictionary KSVD-128, the number of dictionary atoms seems too small, and some image patches (e.g., those with significant details) are not well represented. Therefore, the fusion performance is not comparable to the one obtained by using the dictionaries KSVD-256 and KSVD-512. However, if the number of dictionary atoms is too large, the atoms become too redundant. This will degrade the fusion performance. KSVD-1024 is one such example. In addition, this will also increase the computational complexity of a fusion method.
\begin{table}[h]
  \centering
  \caption{Performance of different dictionaries on the two sets of test images. Scores for all image pairs in each dataset are averaged.}\label{6}
  \begin{tabular}{c | c | c c c c c}
   \toprule
   \makecell{Test\\images}& Dictionary & \emph{MI} & $Q_G$ &$Q_S$ &$Q_{ZP}$ & $Q_{PC}$\\
   \midrule
   \multirow{5}{*}{\makecell{Multi-\\focus\\images}}& DCT-512 &3.9947	&0.7350	&0.4732	&0.8941	&0.6443\\
   & KSVD-128 & 3.8924	&0.7439	&0.4737	&0.9012	&0.6620\\
   & KSVD-256 & 4.0344	&0.7575	&0.5003	&0.9523	&0.6826\\
   & KSVD-512 & \textbf{4.1267}	&\textbf{0.7584}	&\textbf{0.5008}	&\textbf{0.9533}	&\textbf{0.6846}\\
   & KSVD-1024 & 4.0588	&0.7532	&0.4919	&0.9321	&0.6753\\
   \hline
   \multirow{8}{*}{\makecell{Visible-\\infrared\\images}}& DCT-512 &2.3280	&0.5892	&0.3939	&0.8195	&0.4082\\
   & KSVD-128 & 2.0021	&0.5941	&0.4009	&0.7680	&0.3858\\
   & KSVD-256 & 2.2390	&0.6179	&0.4210	&0.8287	&0.4175\\
   & KSVD-512 & \textbf{2.3239}	&\textbf{0.6192}	&\textbf{0.4225}	&0.8340	&0.4208\\
   & KSVD-1024 & 2.2528	&0.6088	&0.4158	&0.8218	&0.4124\\
   & D$_{vi}$-512 & 2.3111	&0.6121	&0.4196	&\textbf{0.8406	}&\textbf{0.4218}\\
   & D$_{ir}$-512 & 2.1703	&0.6051	&0.4126	&0.8106	&0.4059\\
   & D$_{joint}$-512 & 2.2774	&0.6109	&0.4184	&0.8357	&0.4216\\
  \bottomrule
  \end{tabular}
\end{table}

\subsection{Activity levels}
Thereafter, we discuss the impact of three activity level measures, $l_0$-norm, $l_1$-norm and $l_2$-norm of representation coefficients in (29), on the fusion performance. In this experiment, we employ the traditional SR model and the maximum-selecting fusion rule during the fusion process. The quantitative values obtained by the image fusion quality measures considered in Table 9 indicate that the $l_1$-norm of representation coefficients is a better choice among the three activity levels mentioned here. It achieves higher scores for the fusion of multi-focus images as well as for the fusion of multi-modality images, especially for the former.

\begin{table}[h]
  \centering
   \caption{Performance of different activity levels on the two sets of test images. Scores for all image pairs in each dataset are averaged.}\label{7}
  \begin{tabular}{c | c | c c c c c}
   \toprule
   \makecell{Test\\images}& \makecell{Activity\\level} & \emph{MI} & $Q_G$ &$Q_S$ &$Q_{ZP}$ & $Q_{PC}$\\
   \midrule
   \multirow{3}{*}{\makecell{Multi-\\focus\\images}}& $l_0$-norm &\textbf{4.5006}	&0.7098	&0.4774	&0.9761	&0.6529\\
   & $l_1$-norm & 4.1267	&\textbf{0.7584}	&\textbf{0.5008}	&\textbf{0.9533}	&\textbf{0.6846}\\
   & $l_2$-norm & 4.0761	&0.7557	&0.5025	&0.9473	&0.6755\\
   \hline
   \multirow{3}{*}{\makecell{Visible-\\infrared\\images}}&$l_0$-norm &\textbf{2.9882}	&0.5826	&0.4037	&\textbf{0.9679}	&\textbf{0.5246}\\
   & $l_1$-norm & 2.3239	&\textbf{0.6192}	&\textbf{0.4225}	&0.8340	&0.4208\\
   & $l_2$-norm & 2.3390	&0.6135	&0.4217	&0.8453	&0.4242\\
  \bottomrule
  \end{tabular}

\end{table}

\section{Conclusion and Discussion}
SR-based image fusion methods have attracted much attention recently. Sparse representation models, dictionary learning, and fusion rules are three key components of in these techniques. In this paper, we have presented a thorough survey on the issues related to SR-based fusion methods. The following conclusions could be drawn accordingly.

	For representation models, the traditional SR model is the most popular in image fusion. Extensions, such ASR, GSR, NNSR, JSR, and RSR models, have also been applied to image fusion. Fusion performance varies with these models depending on the application. For example, GSR generally achieves better fusion performance when applied to multi-focus image fusion as well as infrared and visible image fusion. RSR and NNSR might also be a good choice for the fusion of multi-focus images and multi-modality images, respectively.

	Regarding the dictionaries, the over-complete dictionaries with a fixed basis (e.g., a DCT basis) and those learned from a set of training images (\emph{global trained dictionary}) or the input images themselves (\emph{adaptively trained dictionary}) have been applied to image fusion. Generally, the learned dictionaries could achieve better fusion performance than those with a fixed basis. The number of atoms in a dictionary has a strong impact on the fusion performance. A compact dictionary with good representation capability is greatly desirable in image fusion for high fusion performance and computational efficiency. However, this is still a challenging problem in that area.

	For fusion strategies, the $l_0$-norm, $l_1$-norm and $l_2$-norm of the representation coefficients or reconstruction errors are usually employed as the activity level. The maximum-selecting fusion rule is employed in most of the existing SR-based image fusion methods. Designing more sophisticated activity levels and fusion rules for SR-based image fusion presents an interesting research topic for the future.

	Moreover, most of the current SR-based fusion methods are performed in a patch-based way. In order to improve the robustness to mis-registration while reducing the spatial artifacts, a sliding window technology is often employed. This results in the loss of information in the fused image and in the huge increase of computational complexity. A good alternative fusion strategy might consist of integrating some local consistency prior into these SR models during the sparse coding phase for each image patch.

	Finally, while we mainly reviewed in this paper SR-based fusion methods that have been applied to multi-focus and multi-modality image fusion, it is also worth noting that the SR theory has also been exploited in some other applications in image fusion, such as remote image fusion (also called pan-sharpening) \cite{guo2014an,zhu2013a,vicinanza2015a,jiang2014two-step} and multi-exposure image fusion \cite{wang2014exposure}. SR-based pan-sharpening is a hot topic in this field.


%



\ifCLASSOPTIONcompsoc
  \section*{Acknowledgments}
\else
  \section*{Acknowledgment}
\fi

    This work is supported by the National Natural Science Foundation of China under Grant No.61104212, by Natural Science Basic Research Plan in Shaanxi Province of China (Program No. 2016JM6008), by the Fundamental Research Funds for the Central Universities under Grant No. NSIY211416, and by the Natural Science Foundation under Grant No. ECCS-1405579.

\ifCLASSOPTIONcaptionsoff
  \newpage
\fi

\bibliographystyle{IEEEtran}
\bibliography{ref}
\end{document}